\pdfoutput=1

\documentclass[11pt]{article}

\usepackage[preprint]{acl}

\usepackage{times}
\usepackage{latexsym}

\usepackage[T1]{fontenc}

\usepackage[utf8]{inputenc}

\usepackage{microtype}

\usepackage{inconsolata}

\usepackage{graphicx}
\usepackage{xcolor}
\usepackage{xspace}
\usepackage{amsmath}
\usepackage{cleveref}
\usepackage{framed}
\usepackage{mdframed}
\usepackage{multirow}
\usepackage{booktabs}
\usepackage{pifont}
\usepackage{float}
\usepackage{color}
\usepackage{multicol}
\usepackage{lscape}
\usepackage{colortbl}
\usepackage{subcaption} 
\usepackage{tabularx}
\usepackage{tikz}
\usetikzlibrary{shapes.geometric, arrows}

\tikzstyle{startstop} = [rectangle, rounded corners, minimum width=3cm, minimum height=1cm,text centered, draw=black, fill=red!30]
\tikzstyle{process} = [rectangle, minimum width=3cm, minimum height=1cm, text centered, draw=black, fill=blue!20]
\tikzstyle{decision} = [diamond, minimum width=3cm, minimum height=1cm, text centered, draw=black, fill=yellow!30]
\tikzstyle{arrow} = [thick,->,>=stealth]

\usepackage{enumitem}
\usepackage{wrapfig}
\usepackage{booktabs}
\usepackage{amsmath, amssymb}
\usepackage{tabularx}

\colorlet{lightblue}{blue!15}
\colorlet{lightorange}{orange!15}
\colorlet{lightgreen}{green!15}
\colorlet{lightgray}{gray!15}
\colorlet{lightyellow}{yellow!15}
\colorlet{lightpink}{pink!30}

\newcommand{\brihi}[1]{{\color{blue} [{\bf Brihi}: #1]}}
\newcommand{\keyu}[1]{{\color{teal} [{\bf Keyu}: #1]}}
\newcommand{\sadra}[1]{{\color{orange} [{\bf Sadra}: #1]}}
\newcommand{\sahana}[1]{{\color{violet} [{\bf Sahana}: #1]}}
\newcommand{\swabha}[1]{{\color{green!70!black} [#1]$_\text{SS}$}}
\newcommand{\xiang}[1]{{\color{red} [{\bf Xiang}: #1]}}
\newcommand{\kaitlyn}[1]{{\color{red!70!black} [{\bf Kaitlyn}: #1]}}

\newcommand{\dataset}{\textsc{ELI-Why}\xspace}
\newcommand{\datasetnum}{13,392\xspace}
\newcommand{\gptfour}{GPT-4\xspace}
\newcommand{\llama}{Llama-3.2-3B-Instruct\xspace}
\newcommand{\qwen}{Qwen 2.5 14B Instruct\xspace}
\newcommand{\deepseek}{DeepSeek R1 Distill LLama 8B\xspace}
\newcommand{\llamaseventy}{Llama-3.3-70B-Instruct\xspace}
\newcommand{\sigoverall}{Overall\xspace}
\newcommand{\sigorder}{Order Preserved\xspace}
\newcommand{\eg}{e.g.\xspace}
\newcommand{\Eg}{E.g.\xspace}

\newcommand{\tesdiff}{TESDiff\xspace}

\newcommand{\escolor}[1]{{\colorbox{lightblue}{#1}}}
\newcommand{\hscolor}[1]{{\colorbox{lightgreen}{#1}}}
\newcommand{\phdcolor}[1]{{\colorbox{lightorange}{#1}}}
\newcommand{\defaultcolor}[1]{{\colorbox{lightgray}{#1}}}
\newcommand{\googlecolor}[1]{{\colorbox{lightyellow}{#1}}}
\newcommand{\smecolor}[1]{{\colorbox{lightpink}{#1}}}

\newcommand{\es}{\escolor{Elementary School}\xspace}
\newcommand{\esshort}{\escolor{Elementary}\xspace}
\newcommand{\hs}{\hscolor{High School}\xspace}
\newcommand{\hsshort}{\hscolor{High School}\xspace}
\newcommand{\phd}{\phdcolor{Graduate School}\xspace}
\newcommand{\phdshort}{\phdcolor{Graduate}\xspace}
\newcommand{\default}{\defaultcolor{Default}\xspace}
\newcommand{\google}{\googlecolor{Web-Retrieved}\xspace}
\newcommand{\sme}{\smecolor{Manually Web-Retrieved}\xspace}
\newcommand{\successrate}{\% Informative Explanations\xspace}
\newcommand{\successaccuracy}{\% Matched Informative Explanations\xspace}

\renewcommand{\brihi}[1]{}
\renewcommand{\keyu}[1]{}
\renewcommand{\sadra}[1]{}
\renewcommand{\kaitlyn}[1]{}
\renewcommand{\xiang}[1]{}
\renewcommand{\swabha}[1]{}
\renewcommand{\sahana}[1]{}

\title{\dataset: Evaluating the Pedagogical Utility of Language Model Explanations}

\DeclareSymbolFont{extraup}{U}{zavm}{m}{n}
\DeclareMathSymbol{\vardiamond}{\mathalpha}{extraup}{87}
\DeclareMathSymbol{\varheart}{\mathalpha}{extraup}{86}
\newcommand{\aspace}{\hspace{1em}}
\newcommand{\stanford}{$^{\vardiamond}$}
\newcommand{\usc}{$^{\varheart}$}

\author{
    Brihi Joshi\thanks{~~Equal contribution.}\usc\aspace 
    Keyu He$^{*}$\usc\aspace
    Sahana Ramnath\usc\aspace
    Sadra Sabouri\usc\aspace
    \textbf{Kaitlyn Zhou}\stanford\aspace\\
    \textbf{Souti Chattopadhyay}\usc\aspace 
    \textbf{Swabha Swayamdipta}\usc\aspace
    \textbf{Xiang Ren}\usc\aspace\\
    \usc University of Southern California \aspace
    \stanford Stanford University \\
    \texttt{\{brihijos, frankhe\}@usc.edu} \\ \\
    \href{https://inklab.usc.edu/eli-why}{\tt{inklab.usc.edu/eli-why}}
}

\begin{document}
\maketitle
\begin{abstract}
Language models today are widely used in education, yet their ability to tailor responses for learners with varied informational needs and knowledge backgrounds remains under-explored. 
To this end, we introduce \dataset, a benchmark of 13.4K ``Why'' questions 
to evaluate the pedagogical capabilities of language models. 
We then conduct two extensive human studies to assess the utility of language model-generated explanatory answers (explanations) on our benchmark, tailored to three distinct educational grades: elementary, high-school and graduate school.
In our first study, human raters assume the role of an ``educator'' to assess model explanations' fit to different educational grades.
We find that \gptfour-generated explanations match their intended educational background only 50\% of the time, compared to 79\% for lay human-curated explanations. 
In our second study, human raters assume the role of a learner to assess if an explanation fits their own informational needs.
Across all educational backgrounds, users deemed \gptfour-generated explanations 20\% less suited on average to their informational needs, when compared to explanations curated by lay people.
Additionally, automated evaluation metrics reveal that explanations generated for different informational needs remain indistinguishable in their grade level, for different language model families, 
limiting their pedagogical effectivenes.
\end{abstract}

\section{Introduction}

\begin{figure}[!t]
    \centering
    \includegraphics[width=\linewidth]{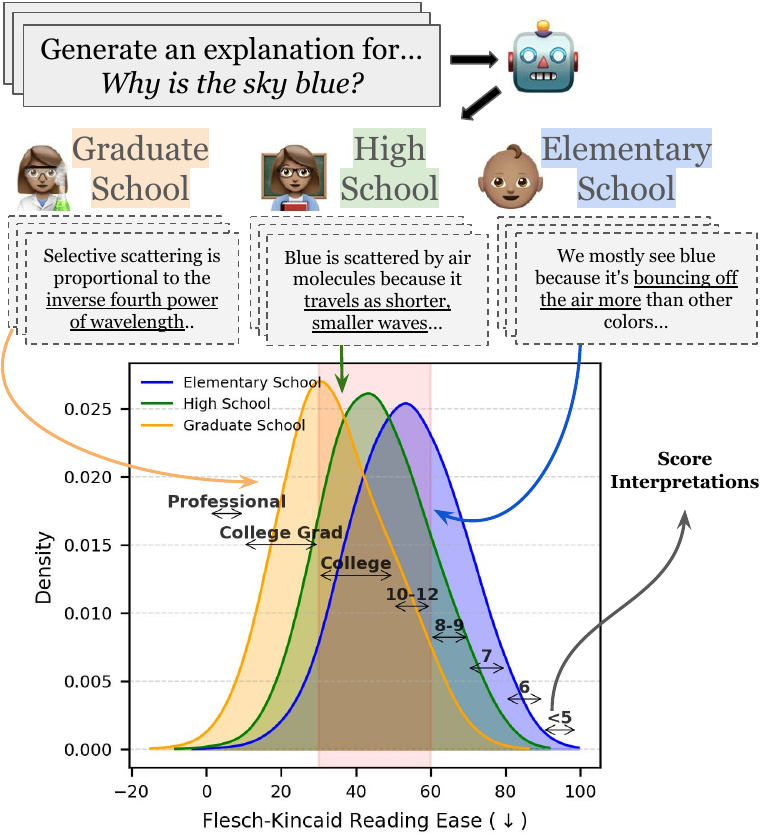}
    \caption{\textbf{Distribution of Flesch-Kincaid Reading Ease scores for tailored \gptfour-explanations in \dataset: }Explanations are generated for elementary, high school, and graduate-level backgrounds for ``Why'' questions. Interpretations of Flesch-Kincaid Reading Ease scores depicting grade-level complexity (lower = more complex) often overlap (within the high school-college range, region highlighted in pink).}
    \label{fig:auto_metrics_interpretation}
\end{figure}

Language models are increasingly used in education to seek information \cite{suri2024use}, tutoring \cite{chevalier2024language}, and automated assessment \cite{tlili2023if,stahl2024exploring}. 
A critical aspect of their pedagogical utility is their potential to \textit{tailor} responses to learners with varying informational needs \cite{adolphe2023cogtraining, puech2024pedagogicalsteeringlargelanguage, avies2021edai, chevalier2024language, jurenka2024responsibledevelopmentgenerativeai, sun2024genaicollege,ross-andreas-2024-toward}. 
This is particularly important in scientific communication, where complex concepts must be conveyed effectively to nonexperts \cite{august2023paperplain}, and in policy or legal communication, where text must balance technical accuracy with readability \cite{cheong2024ailawyerbutengaging}. 
Despite the potential of language models to modify explanations in their complexity~\cite{august2024knowyouraudience}, formality \cite{luo2023prompt}, and domain specificity \cite{karabacak2023embracing,wang2023grammar} at inference time, it remains unclear whether they can effectively generate responses that are useful both to educators~\cite{kim2024designing} and to learners alike~\cite{lee2023dapie}. 

One critical challenge in pedagogy is answering ``Why'' questions.
These require explanatory answers to meet different learners where they are.
For example, for the question ``\textit{Why is the sky blue?}'', a high school student might find the explanation ``\textit{Sunlight scatters when it hits air molecules}'' more understandable, while a physics graduate might find a more technical answer ``\textit{Selective scattering is proportional to the inverse fourth power of wavelength}'' more satisfactory.
Although language models are capable of step-by-step reasoning across various tasks \cite{wei2022chain, prystawski2023think}, by default they generate a \textit{one-size-fits-all} explanation, that might not fit the informational needs of a user interacting with it~\cite{august2024knowyouraudience}. 
Can the prompt-following skills of language models~\cite{wei2021finetuned, zeng2023evaluating, lee2024aligningthousandspreferencesmessage} help them tailor their explanations\footnote{For brevity, we henceforth refer to free-text explanatory answers to ``Why'' questions as \textit{explanations}.} to users with different informational needs?

We introduce \dataset\footnote{Name inspired by the subreddit ``Explain Like I'm Five'', where users seek simpler answers to questions. \url{https://www.reddit.com/r/explainlikeimfive/}}, a dataset of 13.4K ``Why'' questions that span different disciplines such as science, medicine, and humanities, 
such as ``\textit{Why do countries have flags?}'' or ``\textit{Why do leaves change color in the fall?}'' to examine the pedagogical utility of language model explanations. 
While prior studies have explored the ability of language models to generate general-purpose explanations in a pedagogical setting~\cite{joshi-etal-2023-machine, li2024chatgpt}, it is important that explanations adapt to the prior knowledge of learners \cite{schmucker2024ruffle, ye2024storypark, lee2023dapie}. 

Our experimental settings involve using the \textit{highest educational degree attained} as a proxy for the informational needs of a user. 
Specifically, we prompt language models to generate three different explanations to \dataset questions, fit for users with elementary school, high school, or graduate level education.
We conduct automated evaluations and two human studies to assess the utility of language model generated grade-tailored explanations. 
Our first human study is conducted from the perspective of an educator to test the appropriateness of an explanation for users with different educational backgrounds on a subset of \dataset.
We find that \gptfour-generated explanations match their intended background only 50\% of the time, compared to 79\% for explanations curated by lay humans (\S\ref{sec:educator}).
We then use automated metrics to assess the grade-level readability of explanations; while explanations become more lengthy and contain `complex' words as the educational level increases, their complexity in terms of grade-level readability often overlaps (shown in \Cref{fig:auto_metrics_interpretation} and \Cref{sec:auto_metrics}).
We extend this automated metric analysis to three more model families apart from \gptfour, and report similar findings.
Our second human study tests the appropriateness of an explanation from the perspective of a learner's own self-reported informational needs (\Cref{sec:learner}).
To capture the information needs of users, we asked participants to rate the explanations based on whether they provide new information and whether the explanations connect to their prior knowledge.
Studies with participants from elementary, high school, and graduate backgrounds (Physics and Psychology) reveal that \gptfour-generated explanations are relatively 20\%
less informative than explanations curated by lay humans.
This gap is particularly pronounced for users with graduate-level and high-school backgrounds. 


Overall, our results highlight the limitations of current language model-driven pedagogy and suggest that explicitly prompting for audience adaptation alone might be insufficient.
We believe that in addition to \dataset being a valuable resource to evaluate language models' pedagogical utility, our human-centered evaluation framework can help evaluate personalized agents catered to the informational needs of individual users.

\section{
The \dataset Benchmark}
\label{sec:dataset}
Existing work in pedagogical evaluation of language models has either focused on objective benchmark-driven question-answering tasks (\eg multiple-choice science-based question answering) \cite{lu2022learn, mitra2024retllm, chang2025chatbenchstaticbenchmarkshumanai} or subjective use-case driven tasks (\eg evaluating academic achievements induced by language model assistants) \cite{hoper2024new, sun2024genaicollege}.
Combining these two, we focus on the task of answering 
``Why'' questions; they ensure a good balance between having a knowledge-seeking setting and having room for subjectivity in the manner in which knowledge is presented \cite{sulik2023explsinthewild}.
To this end, we introduce \textbf{\dataset}, which consists of \datasetnum ``Why'' questions curated across STEM and Non-STEM domains.
There are 6,217 STEM questions (across disciplines like Physics, Chemistry, Computer Science, Material Engineering etc.), and 7,175 non-STEM questions (across disciplines like Sociology, Law, Culture, History, Public Relations, etc.). 
Our dataset is created by (1) \textbf{over-generating} ``Why'' questions from \gptfour via few-shot prompting
followed by (2) \textbf{extensive filtering} by checking
validity of the generated questions. 
We expand upon these steps below, and provide full details, including prompts, model settings and filtering process about \dataset curation in \Cref{sec:apx:dataset}.

\paragraph{Overgenerating ``Why'' questions from \gptfour.}   
We use a set of 50 seed ``Why'' questions from \citet{sulik2023explsinthewild} (\Cref{tab:seed questions}) and split them into different disciplines.
We use a random subset of these 
as in-context examples to prompt \gptfour (\Cref{tab:question_generation_config_prompt}) to generate more questions in a given discipline \cite{liu-etal-2022-wanli}.
This led to a set of $\sim$30k questions.

\paragraph{Filtering generated questions.}
We then manually deduplicated questions 
from the set.
We additionally removed niche, domain-specific questions (\eg questions like ``\textit{Why is the electron cloud model currently the most accepted atomic model?}'') with the help of crowdworkers.
Details about the filtering process can be found in \Cref{sec:apx:question_filter}.
This resulted in the final \datasetnum questions.

\section{Generating Explanations for different Educational Backgrounds}
\label{sec:generating_explanations}
Users with varying educational or conceptual backgrounds differ in expectations of answers to their questions~\cite{kolb2007kolb, bertrand2023selective}. 
Tailoring responses to users with different educational backgrounds is important to improve language models' use in pedagogy~\cite{adolphe2023cogtraining, puech2024pedagogicalsteeringlargelanguage}.
In this section, we describe the different educational levels we used for evaluating language model explanations and our methodology for generating grade-tailored explanations. 

\paragraph{Educational backgrounds.} 
We choose three educational levels with different informational needs\footnote{As educational levels can often overlap with informational needs, we choose three levels that are ideally least overlapping.} for our users\footnote{Throughout this study, we refer to adults aged 18+ as our user base.}: \es, \hs and \phd, in the context of education in the United States\footnote{\url{https://usahello.org/education/children/grade-levels/}}.
\es group typically covers content up to U.S. Grade 4, and adults with this education level may have limited theoretical knowledge of individual disciplines.
The \hs group extends through U.S. Grade 12 to approximately the sophomore year of undergraduate studies, and adults at this level have a foundational grasp of academic subjects but may still struggle with discipline-specific terminology.
\phd group typically follow a bachelor's degree, offering advanced, specialized education and adults with this education have few knowledge gaps and possess expertise in specific areas without needing foundational instruction\footnote{Description of these backgrounds are informed by \url{https://en.wikipedia.org/wiki/Educational_attainment_in_the_United_States} and~\citet{falk2013factors}.}.

\paragraph{Generating grade-tailored explanations.}
For any given ``Why'' question, our goal is to generate three responses corresponding to users whose highest educational degree is at the \es, \hs and \phd level.
We generate explanations for each question by zero-shot prompting language models from four model families---\gptfour-0613\footnote{Last accessed August 2024.} (henceforth shortened to \gptfour), \llama, \qwen and \deepseek.	
We instruct each language model to assume the role of an \textit{expert} in order to provide suitable explanations for each of the three educational backgrounds (prompt detailed in \Cref{sec:apx:expl_generation}).\footnote{Language models like \gptfour contains knowledge from varied disciplines, as per \url{https://openai.com/index/gpt-4-research/}.} 
Additionally, our prompts contain instructions like ``\textit{do not add any additional text like greetings or ornamental words}'' to ensure that language models tailors the response in terms of knowledge and not just stylistic cues.
For example, \gptfour would often add context, such as ``\textit{playing in the park}'' or ``\textit{other kids}'' while generating explanations for \es background. 
We try to limit such generations using specific instructions in the prompt, so that it puts less emphasis on stylistic verbiage, when compared to knowledge content (\Cref{tab:prompts_used}).
Throughout the rest of this paper, we use \textit{intended educational background} of an explanation to refer to the educational background used to generate the explanation.
All model parameters used to generate explanations are detailed in \Cref{sec:apx:model_details}. 
While all four language models are used for automated evaluations, we only use explanations generated by \gptfour for our human studies.
\paragraph{Baseline explanations.}
In addition to the above explanations, we prompt language models to produce the \default explanation for a given question, without providing any educational background (prompt detailed in \Cref{sec:apx:expl_generation}).
We also collect baseline \google explanations using the Google API\footnote{\url{https://serpapi.com/}}; we use the \textit{Featured Snippet} provided by Google.\footnote{These are pre-Gemini summary results, where the only use of a model had been to rank relevant snippets, according to \url{https://support.google.com/websearch/answer/9351707?hl=en.}}


\paragraph{Web Explanations Curated by Lay Humans.}
Lastly, for a subset of 40 questions in \dataset, authors of this work manually curated explanations (\sme) for each educational background, by searching appropriate websites.
All explanations are curated independently by two authors, then discussed together to preserve the most plausible explanation.
For \eg, we retrieve \phd level explanations for a question by searching through journals and research papers on the topic, and \es level explanations by searching through the Explain Like I'm Five (ELI5) subreddit\footnote{\url{https://www.reddit.com/r/explainlikeimfive/}}.
For \hs, we retrieve explanations from blog posts and web pages intended for lay users.
These are not meant to be expert-level explanations, but simulate a process of obtaining explanations for different grade levels in contrast to language model generations \cite{oh2008use, ward2021people}.

\section{Do language model explanations match their intended educational background?}
\label{sec:educator}
In this section, we evaluate whether grade-tailored language model explanations match their intended educational backgrounds, using human evaluations.
We then extend to a large-scale empirical analysis on all of \dataset and model variants, where we employ different automated metrics and reconcile these findings with that of the user study.

\subsection{Intended vs. perceived educational backgrounds of tailored explanations}
\label{sec:perceived:human_eval}

\begin{figure}[!t]
    \centering
    \includegraphics[width=0.8\columnwidth]{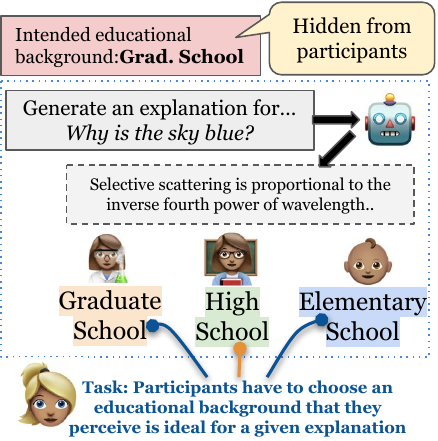}
    \caption{\textbf{User study for evaluating perceived background match:} 
    Participants assume the role of an educator and determine the educational background of the explanation presented to them, without knowing the explanation's intended educational background. In this example, the explanation is generated for \phd, but participants perceive it to be ideal for \hs users.
    }
    \label{fig:perceived_background_match_task}
\end{figure}
We define the \textit{intended educational background} as the grade-level for which an explanation was generated. 
We then define the \textit{perceived educational background} as the grade-level that \textit{a human user} associates with an explanation. 
To identify if language model explanations are successfully tailored for different grade levels, we conduct a user study in which participants assume the role of an educator; they read questions and a language model explanation to indicate their perceived educational background of the explanation.
We then evaluate the percentage of explanations where the intended educational background matches the perceived educational background. 
We term this as \textit{Perceived Background Match}.
This formulation allows us to directly measure whether tailored explanations match the grade level they were generated for.

\paragraph{User study design.}
We conducted a user study with a subset of $400$ ``Why'' questions from \dataset, along with explanations generated by \gptfour tailored for each of the three grade levels we consider.
The participants were presented with a question-explanation pair and were asked to identify the perceived educational background of the explanation (\Cref{fig:perceived_background_match_task}). 
Before making their judgments, the participants received detailed task instructions, including information on different educational backgrounds defined in \Cref{sec:generating_explanations}.
Additionally, pilot evaluations, conducted by the authors and a subset of participants, helped refine instruction clarity. 
Each participant annotated five question-explanation pairs, and each pair received three independent annotations, ensuring a diverse evaluation of perceived backgrounds; we considered a majority vote of the perceived educational background for all explanations.
As a control, we also conducted a user study on \sme explanations for 40 questions to understand perceived explanation match trends for explanations curated by lay users.
Further details on participant screening, demographics, and study setup are provided in \Cref{sec:apx:human_exp}.


\paragraph{Results.}
\begin{figure}
    \centering
    \includegraphics[width=0.8\columnwidth]{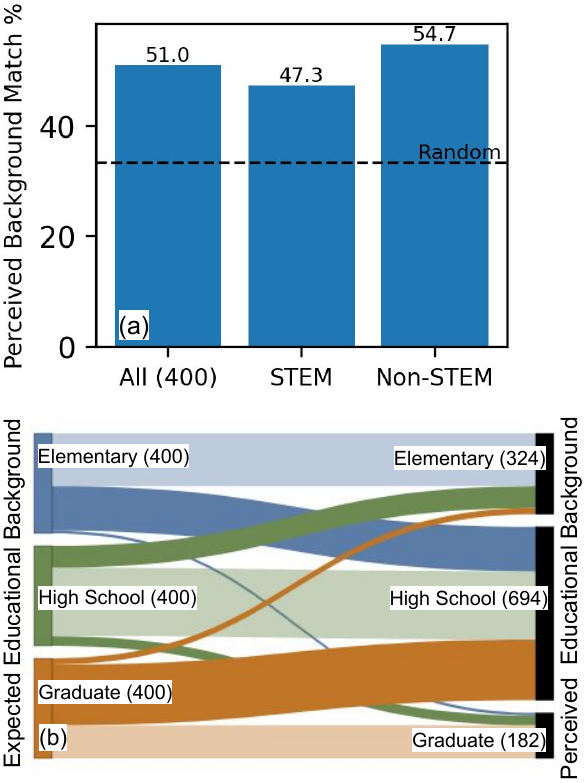}
    \caption{\textbf{Perceived background match results:} (a) Perceived background match \% vs. domain split of questions. (b) Sankey diagram depicting change between intended background of explanations (left) to perceived background (right) after user study.}
    \label{fig:perceived_background_match_results}
\end{figure}
\Cref{fig:perceived_background_match_results} presents results from the user study.
\Cref{fig:perceived_background_match_results}(a) shows that the perceived background match \% of tailored explanations generated by \gptfour is very low (close to 50\%).
This trend is observed across STEM and Non-STEM splits of the subset.
Furthermore, the user study also reveals that tailoring \textit{mismatch} is seen across the board for all educational backgrounds. 
\Cref{fig:perceived_background_match_results}(b) shows the change between intended and perceived educational background, after the study.
Most explanations are perceived to be tailored for \hs, which can be explained by \gptfour's tendency to be conditioned towards a ``lay-user'' \cite{august2024knowyouraudience, hsu2024free}.
We also observe surprising mismatches---\eg. \es explanations being perceived as \phd, and vice versa.
We show examples of these cases, along with justification written by users in \Cref{sec:apx:anno2.1}.
Additionally, the perceived role match of \sme explanations is much higher ($79.16\%$). 
This reveals a concerning trend in \gptfour's explanations: while \gptfour can be easily prompted to generate explanations tailored for different educational backgrounds, it does not necessarily mean that users perceive these explanations fit for a given background, potentially hindering \gptfour's utility in pedagogy \cite{kasneci2023chatgpt}.

\subsection{What do automated metrics reveal about tailored language model explanations?}
\label{sec:auto_metrics}

\begin{table*}[t]
\centering
\resizebox{\linewidth}{!}{
\begin{tabular}{lllccccc}
\toprule
\textbf{Type} & \textbf{Metric} & \textbf{Language Model} & \es & \hs & \phd & \default & \google \\
\midrule
\multirow{17}{*}{Surface-form} & \multirow{4}{*}{\# Sentences} 
& \gptfour 
& 04.63 \(\pm\) \small{01.34}
& 07.08 \(\pm\) \small{02.53}
& 08.46 \(\pm\) \small{02.62}
& 05.07 \(\pm\) \small{01.63} 
& \multirow{4}{*}{02.30 \(\pm\) \small{00.90}} \\
& & \llama & 03.29 $\pm$ \small{01.63} & 06.70 $\pm$ \small{02.97} & 09.10 $\pm$ \small{03.33} & 04.24 $\pm$ \small{02.63} & \\
& & \qwen & 03.38 $\pm$ \small{00.86} & 03.77 $\pm$ \small{00.93} & 04.50 $\pm$ \small{01.17} & 02.97 $\pm$ \small{00.87} & \\
& & \deepseek & 04.45 $\pm$ \small{02.60} & 05.30 $\pm$ \small{02.88} & 06.50 $\pm$ \small{03.33} & 04.78 $\pm$ \small{02.93} & \\
\cmidrule{2-8}

 & \multirow{4}{*}{\shortstack[l]{Avg. \# Words \\ / Sentence}} 
& \gptfour 
& 18.43 \(\pm\) \small{03.47}
& 19.17 \(\pm\) \small{03.36}
& 20.00 \(\pm\) \small{03.38}
& 19.35 \(\pm\) \small{03.57}
& \multirow{4}{*}{17.26 \(\pm\) \small{06.90}} \\
& & \llama & 20.39 $\pm$ \small{05.10} & 21.30 $\pm$ \small{03.81} & 23.12 $\pm$ \small{03.69} & 23.74 $\pm$ \small{04.88} & \\
& & \qwen & 18.50 $\pm$ \small{03.75} & 19.92 $\pm$ \small{03.74} & 21.73 $\pm$ \small{03.96} & 21.54 $\pm$ \small{04.49} & \\
& & \deepseek & 20.22 $\pm$ \small{04.40} & 19.89 $\pm$ \small{03.97} & 20.03 $\pm$ \small{03.69} & 20.47 $\pm$ \small{04.42} & \\
\cmidrule{2-8}

 & \multirow{4}{*}{\textbf{\shortstack[l]{Avg. Reading\\ Time (s)*}}} 
& \gptfour           & 06.36 $\pm$ \small{01.75} & 10.57 $\pm$ \small{03.65} & 13.93 $\pm$ \small{04.05} & 07.81 $\pm$ \small{02.41} & \multirow{4}{*}{02.93 $\pm$ \small{01.04}} \\
& & \llama & 04.61 $\pm$ \small{02.14} & 10.97 $\pm$ \small{05.00} & 17.05 $\pm$ \small{06.30} & 07.93 $\pm$ \small{04.71} & \\
& & \qwen & 04.60 $\pm$ \small{01.17} & 05.93 $\pm$ \small{01.41} & 08.08 $\pm$ \small{01.98} & 05.19 $\pm$ \small{01.40} & \\
& & \deepseek & 07.37 $\pm$ \small{04.38} & 08.77 $\pm$ \small{04.97} & 11.22 $\pm$ \small{06.10} & 08.15 $\pm$ \small{05.08} & \\
\cmidrule{2-8}

 & \multirow{4}{*}{\textbf{TE Score}* } 
& \gptfour & 00.43 $\pm$ \small{00.09} & 00.49 $\pm$ \small{00.09} & 00.55 $\pm$ \small{00.09} & 00.50 $\pm$ \small{00.09} & \multirow{4}{*}{00.44 $\pm$ \small{00.12}} \\
& & \llama & 00.37 $\pm$ \small{00.11} & 00.47 $\pm$ \small{00.09} & 00.54 $\pm$ \small{00.10} & 00.53 $\pm$ \small{00.11} & \\
& & \qwen & 00.38 $\pm$ \small{00.10} & 00.47 $\pm$ \small{00.09} & 00.53 $\pm$ \small{00.10} & 00.51 $\pm$ \small{00.10} & \\
& & \deepseek & 00.58 $\pm$ \small{00.14} & 00.60 $\pm$ \small{00.14} & 00.66 $\pm$ \small{00.16} & 00.62 $\pm$ \small{00.14} & \\
\midrule

\multirow{13}{*}{Readability} & \multirow{4}{*}{\textbf{\shortstack[l]{Flesch-Kincaid \\ Reading Ease*} ($\downarrow$)}} 
& \gptfour & 53.82 $\pm$ \small{14.52} & 45.51 $\pm$ \small{14.26} & 34.70 $\pm$ \small{14.43} & 41.00 $\pm$ \small{15.73} & \multirow{4}{*}{53.35 $\pm$ \small{18.52}} \\
& & \llama & 60.91 $\pm$ \small{17.14} & 46.39 $\pm$ \small{15.11} & 34.56 $\pm$ \small{14.51} & 33.68 $\pm$ \small{16.87} & \\
& & \qwen & 57.59 $\pm$ \small{15.19} & 42.38 $\pm$ \small{16.05} & 30.67 $\pm$ \small{16.18} & 33.79 $\pm$ \small{17.44} & \\
& & \deepseek & 34.40 $\pm$ \small{16.79} & 34.59 $\pm$ \small{15.70} & 30.98 $\pm$ \small{15.23} & 30.94 $\pm$ \small{16.69} & \\
\cmidrule{2-8}

 & \multirow{4}{*}{\shortstack[l]{Linsear Write \\ Formula ($\uparrow$)}} 
& \gptfour & 11.67 $\pm$ \small{02.77} & 12.15 $\pm$ \small{02.70} & 13.21 $\pm$ \small{02.70} & 13.16 $\pm$ \small{02.82} & \multirow{4}{*}{11.10 $\pm$ \small{05.21}} \\
& & \llama & 12.29 $\pm$ \small{03.94} & 13.23 $\pm$ \small{03.09} & 14.88 $\pm$ \small{02.93} & 16.43 $\pm$ \small{03.93} & \\
& & \qwen & 11.46 $\pm$ \small{03.02} & 13.61 $\pm$ \small{02.95} & 15.40 $\pm$ \small{02.94} & 15.44 $\pm$ \small{03.49} & \\
& & \deepseek & 14.42 $\pm$ \small{03.70} & 14.03 $\pm$ \small{03.37} & 14.20 $\pm$ \small{03.17} & 14.86 $\pm$ \small{03.79} & \\
\cmidrule{2-8}

& \multirow{4}{*}{\shortstack[l]{Dale-Chall \\ Readability Score ($\uparrow$)}} 
& \gptfour & 09.31 $\pm$ \small{01.16} & 09.84 $\pm$ \small{01.10} & 10.55 $\pm$ \small{01.10} & 10.35 $\pm$ \small{01.23} & \multirow{4}{*}{09.94 $\pm$ \small{01.56}} \\
& & \llama & 08.73 $\pm$ \small{01.47} & 09.49 $\pm$ \small{01.19} & 09.88 $\pm$ \small{01.13} & 10.80 $\pm$ \small{01.38} & \\
& & \qwen & 09.01 $\pm$ \small{01.32} & 10.32 $\pm$ \small{01.32} & 10.91 $\pm$ \small{01.27} & 11.11 $\pm$ \small{01.41} & \\
& & \deepseek & 11.09 $\pm$ \small{01.35} & 11.02 $\pm$ \small{01.25} & 11.14 $\pm$ \small{01.20} & 11.36 $\pm$ \small{01.36} & \\
\midrule

\multirow{4}{*}{Reasoning} & \multirow{4}{*}{\shortstack[l]{\% Mechanistic \\Reasoning}} 
& \gptfour & 54.76\% & 57.28\% & 63.54\% & 58.51\% & \multirow{4}{*}{65.16\%} \\
& & \llama & 55.04\% & 57.81\% & 65.65\% & 63.03\% & \\
& & \qwen & 52.75\% & 57.42\% & 63.61\% & 60.53\% & \\
& & \deepseek & 56.72\% & 56.14\% & 58.48\% & 57.15\% & \\
\bottomrule
\end{tabular}
}
\caption{\textbf{Comparison of surface-form, readability, and reasoning-type metrics across different education levels, along with retrieved explanations.} * represents metrics that have high correlation with user evaluations of perceived educational backgrounds. $\uparrow$ and $\downarrow$ depict direction of scores representing more complex explanations for readability metrics; for all the other metrics, higher values indicate higher complexity.
}
\label{tab:automated metrics}
\end{table*}

\Cref{sec:perceived:human_eval} demonstrated that \gptfour-generated rationales often mismatch their intended educational backgrounds. 
We extend the scale of our analysis to the full \dataset benchmark and more language model families using automated metrics and show that careful interpretation of these metrics also highlight the above mismatch.

\paragraph{Automated Metrics.}

We use three categories of automated metrics, based on surface-form features, readability, and reasoning styles to evaluate whether these automated metrics distinguish between explanations tailored to different grades.
Surface form metrics compute sentence count, average sentence length, estimated reading time \cite{demberg2008data}, and TE Score \cite{august2024knowyouraudience} (the TE score / Thing Explainer Out-of-Vocabulary score measures the proportion of `complex words' in an explanation by taking the proportion of words outside a curated list of the 2,000 most common English words).
We employ three popular readability metrics: Flesch-Kincaid Reading Ease \cite{Flesch1948-gk}, Linsear Write Formula \cite{o1966gobbledygook}, and Dale-Chall Readability Score \cite{dale1948formula}.
Each of these metrics also map score ranges to an interpreted U.S. grade level~\cite{Kincaid1975DerivationON}.
Score range mappings for each metric are detailed in ~\Cref{tab:readability_interpretation}.
Finally, we analyze the type of reasoning in the explanations: whether they are mechanistic (describe \textit{how} a phenomenon occurs, \eg \textit{pollen shedding occurs because of desiccation of anther tips}) vs. functional (the \textit{purpose} why a phenomenon occurs, \eg \textit{pollen shedding occurs to facilitate reproduction}) \cite{sulik2023explsinthewild}.
Further details on the calculation of these metrics are in \Cref{sec:apx:auto_metrics}.

\paragraph{Automated metrics reveal that tailored explanations suffer from \textit{interpretation collapse}.}
\Cref{tab:automated metrics} presents average and standard deviation of automated metrics for grade-tailored explanations, along with two baseline explanations: \default and \google.
Across all language models, we can observe that the surface form metrics, specifically number of sentences, differ significantly across different educational levels. 
Particularly, generated explanations get lengthier as the educational level increases.
All models also end up using more `complex words' with increasing educational levels, as shown by the increasing TE Score for all models.
Additionally, all models end up using more mechanistic reasoning and less teleological reasoning as educational levels increase; prior work has often shown that young children often endorse more teleological explanations~\cite{Schachner2017-gy}, also demonstrated here.

\default explanations mimic \hs explanations in all metrics, indicating that explanations generated by \gptfour without any grade-level tailoring are often intended for a \hs user.
On the other hand, \google explanations are more concise than other explanations, but their complexity varies widely, shown by the high standard deviation for all readability tests.
In \Cref{sec:apx:similarity}, we also compare different grade-tailored explanations with \default and \google in terms of informational overlap between explanations.

We observe an interesting pattern demonstrated by the readability metrics.
Consider the Flesch-Kincaid Reading Ease metric (where a lower score indicates higher grade-level readability of a given text).
This is also one of three metrics (among Avg. Reading Time and TE Score) that correlate significantly with user perceived educational levels that we obtain in \Cref{sec:perceived:human_eval} (\Cref{sec:apx:corr_ratio}). 
For all models except \deepseek, we observe that the Flesch-Kincaid Reading Ease metrics are relatively distinct for different educational backgrounds.
However, it is interesting to see that these values are so close to each other that they often fall under the same interpreted U.S. grade level.
For example, for \gptfour explanations we show the Flesch-Kincaid Reading Ease distributions for grade-tailored explanations in~\Cref{fig:auto_metrics_interpretation}.
When these scores are mapped to their \textbf{interpreted U.S. grade levels}, the distributions collapse into a narrow range, primarily between high school and college-level readability.
We term this as \textbf{\textit{interpretation collapse}}, which is also observed for all language models (\Cref{sec:apx:auto_score_dist}).
In fact, for \deepseek, readability score distributions are almost overlapping for all educational levels. 
This is supported by our observations in~\Cref{sec:perceived:human_eval}, where participants often perceive most explanations as tailored for \hs.
The fact that explanations meant for vastly different backgrounds fall into overlapping score ranges suggests that grade-tailored explanations are not meaningfully differentiating at an interpretive level, even if surface form qualities like length and complexity of words increases.
We suggest that automated metrics (like Flesch-Kincaid Reading Ease) to some extent can also be used to measure whether language model explanations are truly tailored to their intended educational backgrounds, provided they are carefully inspected with their corresponding grade-level interpretations.

\subsection{Case Study: Why is there a mismatch between intended and perceived educational backgrounds of tailored explanations?}
As seen in \Cref{fig:perceived_background_match_results}(b), we observe surprising mismatches while tailoring explanations to different educational backgrounds---
particularly where explanations tailored for higher educational levels like \hs or \phd are instead perceived as \es. 
We hypothesize that such mismatches arise because of certain questions being always associated with a particular educational background, hindering \gptfour's (and possibly other language models') ability to generalize for a different educational background.

\begin{figure} 
\centering \includegraphics[width=0.9\columnwidth]{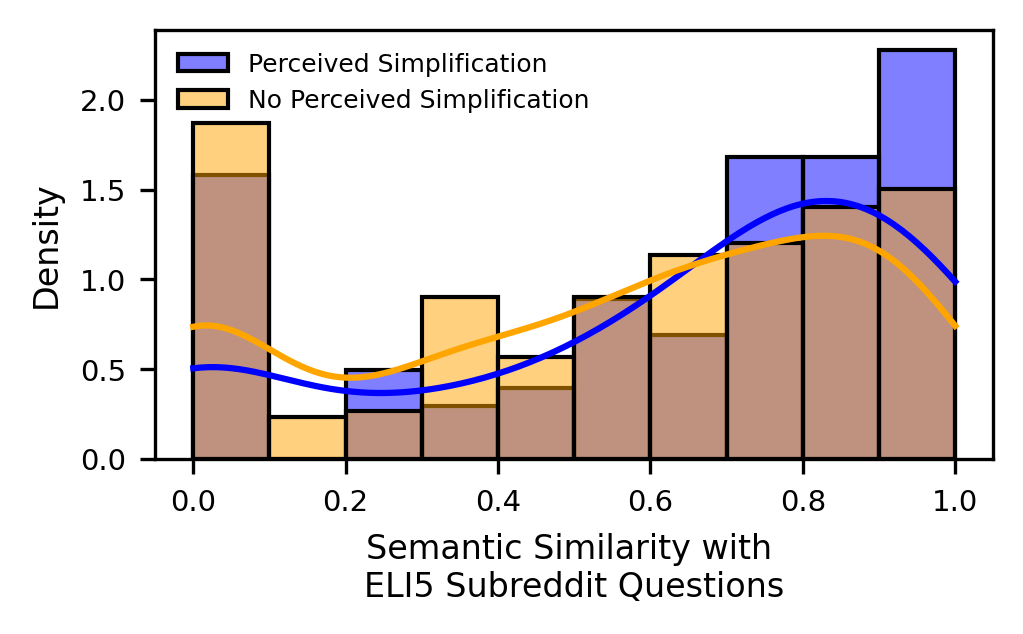} 
\caption{\textbf{Relationship between perceived simplification and semantic similarity to ELI5 questions:} Questions where explanations were perceived as significantly simpler than intended (e.g., intended \hs or \phd but perceived as \es) tend to have higher similarity to questions present the ELI5 subreddit.} 
\label{fig:eli5_simplification} 
\end{figure}

As a case study, we look at the ELI5 subreddit, where users often seek simplified explanations for different questions, most of them being ``Why'' questions (\Cref{sec:apx:case_study_eli5}).
We observe that questions that exhibit perceived simplification---\gptfour's explanations tailored for \hs and \phd that were perceived to be \es by users---are significantly more similar to questions in the ELI5 subreddit than other questions ($p < 0.05$, Mann-Whitney U Test \cite{mann1947test}).
This suggests that \gptfour may overgeneralize and produce simpler explanations when a question closely resembles those always present in contexts pertaining to these educational backgrounds (\Cref{fig:eli5_simplification}).
\section{Do generated explanations help provide \textit{new information} to users?}
\label{sec:learner}

A fundamental notion of utility for language models in pedagogical cases is how much they assist users in learning \textit{new} information \cite{joshi-etal-2023-machine,zhang2024mathemyths,schmucker2024ruffle,lee2023dapie}.
In this section, we discuss the utility of explanations in delivering new information to a \textit{learner}, that aligns with the learner's informational needs.
Understanding this is crucial in determining whether language models like \gptfour tailor explanations for different educational backgrounds merely stylistically or if they provide \textit{new} and \textit{relevant} information that contributes to learning and comprehension.

\paragraph{Evaluating \textit{informativeness} w.r.t user informational needs.}

\begin{figure}
    \centering
    \includegraphics[width=0.73\columnwidth]{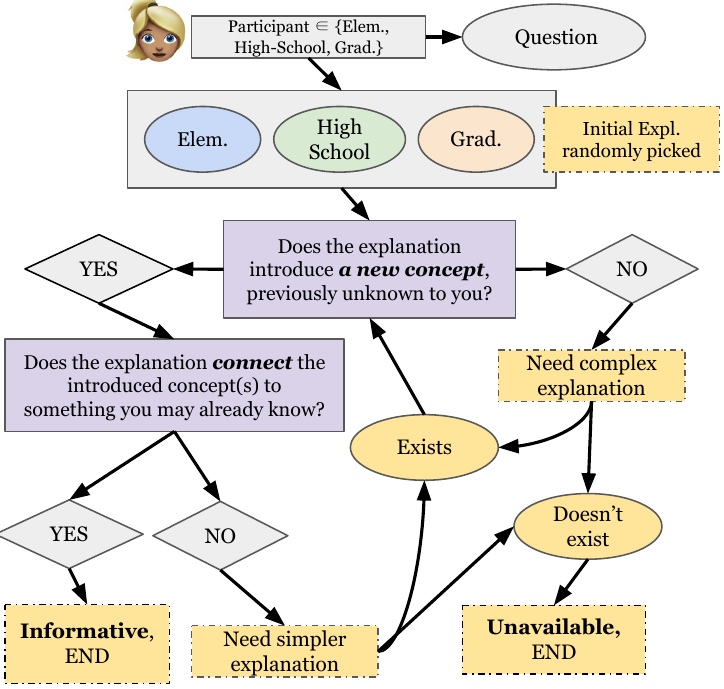}
    \caption{\textbf{User evaluation of explanation informativeness:} Participants are provided with a randomly selected explanation (from one of the educational backgrounds) for a given question. They then assume the role of a learner and determine if an explanation provides new information that connects with their information needs.}
    \label{fig:informativeness_user_evaluation}
\end{figure}

Consider a user with a high school-level background in physics, familiar with basic concepts about light such as such as scattering, wave-particle duality, and light interactions. 
Given a question, ``\textit{Why is the sky blue?}'', the user receives the following explanation: ``\textit{Because of a solar zenith angle (SZA) of 90°, only 1/3 of the blue color of the sky at the zenith is caused by Rayleigh scattering.}''  
While this explanation introduces new terms like \textit{solar zenith angle}, it fails to properly define them, making it difficult for the user to integrate the explanation into their existing knowledge. 
Conversely, an overly simplistic explanation such as \textit{``When sunlight comes through the air bubble that surrounds the Earth, it sometimes hits little bits of air and gets scattered''} provides no meaningful new insights and is therefore uninformative.  

We define that an explanation is \textit{informative} in for a user if it satisfies two conditions: (1) it introduces \textit{new concepts} that the user was previously unaware of, and (2) these new concepts \textit{connect well} with the user’s existing background knowledge, making them easier to understand.
We design the following user study to evaluate the informativeness of an explanation for a given user.
We recruit users belonging to a specific educational background.
The user is presented with a question and a randomly selected stimuli explanation, that could belong to \es, \hs or \phd backgrounds with equal probability.  
The user is then asked: ``\textit{Does the explanation introduce a new concept, previously unknown to you?}''
If the user responds negatively, this implies that the explanation is too simple for them, so the system provides an explanation from the next higher educational background. 
If the user responds positively, they are asked a follow-up question: ``\textit{Does the explanation connect the introduced concept(s) to something you may already know?}''
If they confirm that the concepts are well-integrated, the explanation is considered \textit{informative}. 
However, if the new concepts do not align with their prior knowledge, the explanation introduces new information but lacks coherence, making it difficult for the user to integrate into their understanding; in this case, the system provides an explanation from the next lower educational background. 
\Cref{fig:informativeness_user_evaluation} summarizes this evaluation.

\paragraph{Human Study and Metrics.}

\begin{figure}
    \centering
    \includegraphics[width=0.95\columnwidth]{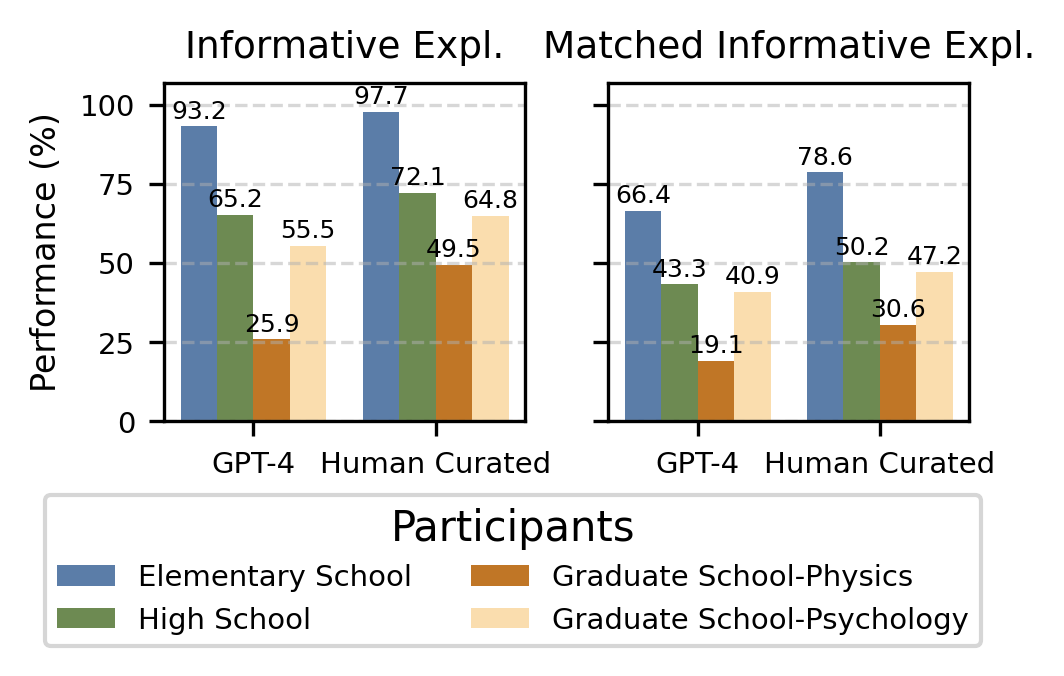}
    \caption{\textbf{Comparison of \successrate and \successaccuracy for across different educational backgrounds.} \gptfour grade-tailored explanations are often informative for \es participants; they struggle to align with the needs of \hs and \phd participants, whereas \sme grade-tailored explanations perform consistently better across all participants.}
    \label{fig:informativeness_results}
\end{figure}

We recruited adult participants with the following highest education levels: 
elementary school,
high school, and graduate degrees in two distinct disciplines—Physics for STEM and Psychology for Non-STEM. 
We select 40 questions from \dataset, and derived from~\Cref{sec:perceived:human_eval}, we use \gptfour-generated explanations that were perceived to match their intended educational backgrounds.
For each educational background and question, participants assume the role of a learner and determine if \gptfour generates explanations that are \textit{informative} for a question. 
Each question is answered by five participants, leading to 200 responses for each educational background.
We compute two metrics: \successrate which is the \% of questions where \textit{any} one of the three \gptfour grade-tailored explanations were found informative,
and \successaccuracy which is the \% of questions where explanations were informative and \textit{matched} the participant's educational background.
Given that we aim to capture how useful grade-tailored explanations are for an individual and that every individual may have different prior knowledge even within the same educational background, we do not do any majority voting while aggregating the above metrics for a question across participants.
Lastly, we also replicate the same user study with the 40 \dataset questions with \sme explanations.

\paragraph{Results.}

\Cref{fig:informativeness_results} shows the \successrate and \successaccuracy results for participants with different educational background.
We observe that participants with higher education backgrounds have lower \successrate and \successaccuracy.
It is particularly stark for participants with a \phd-Physics background, where only 19\% of questions have informative explanations that match the participant's background. 
We find that on an aggregate basis, \sme explanations consistently outperform \gptfour on both metrics across all educational backgrounds. 
While \gptfour provides new information at a comparable rate for \es and \hs participants, its effectiveness declines significantly for \phd-background participants.
On average, for all three educational backgrounds, \sme explanations are relatively 20\% more informative than \gptfour explanations.
It is important to note that \sme explanations are curated by lay experts, not domain experts.
These individuals rely on general knowledge, metadata about online resources to craft responses, yet they still provide more informative and better-aligned explanations than \gptfour. 
This suggests that \gptfour struggles not just with domain expertise, but also with the broader research and adaptation strategies that even nonexperts employ when tailoring explanations. 
Recruiting actual subject matter experts could further widen this gap, highlighting limitations in delivering truly audience-appropriate information.

\section{Conclusion}
Our study introduces \dataset, a benchmark for evaluating the pedagogical utility of language model explanations tailored to users belonging to different educational backgrounds. 
Through user studies, we find that language models like \gptfour struggle to align explanations with intended educational levels and fail to be informative, especially for advanced learners.
Automated evaluations across multiple model families confirm that grade-tailored explanations often collapse into a similar and narrow complexity range, hinting at their limited effectiveness.
Future work can explore methods for integrating measures of pedagogical utility for both educators and learners, as a signal to improve language models for users and cater to more personal learning goals, not bounded by educational backgrounds.
\section*{Acknowledgments}
We are grateful to members of the USC NLP Group, INK Lab, DILL Lab and Aditya Chetan, as well as anonymous reviewers and meta reviewers for their comments and feedback on the draft.
This research is supported in part by the National Science Foundation (NSF) under grant IIS2403437, the Simons Foundation, and the Allen Institute for AI, the Office of the Director of National Intelligence (ODNI), Intelligence Advanced Research Projects Activity (IARPA), via the HIATUS Program contract \#2022-22072200006. 
The views and conclusions contained herein are those of the authors and should not be interpreted as necessarily representing the official policies, either expressed or implied, of NSF, ODNI, IARPA, or the U.S. Government. The U.S. Government is authorized to reproduce and distribute reprints for governmental purposes notwithstanding any copyright annotation therein.
B. Joshi was supported by the Apple Scholars in AI/ML PhD Fellowship.
This work was partially done when S. Swayamdipta was a visitor at the Simons Institute for the Theory of Computing.

\section*{Limitations}
\paragraph{Prompting.} Our evaluations rely exclusively on zero-shot prompting to grade-tailor explanations for all language models, without exploring alternative prompting strategies such as retrieval augmentation or fine-tuning, which may improve such tailoring. 
Lay users of language models often provide prompt-level instructions without additional strategies, which led to our design decision.
Additionally, user interactions are often multi-turn in nature, which we haven't explored in this work.

\paragraph{Human Evaluations.}
Our evaluations are conducted in a controlled setting, where explanations are assessed in isolation rather than within real-world interactive learning contexts. In practice, learners might seek clarification, ask follow-up questions, or engage in dialogue, which could impact how explanations are understood and used~\cite{sulik2023explsinthewild,zhao2024wildchat1mchatgptinteraction}.

\paragraph{Benchmark Design. }
\dataset questions are generated from \gptfour.
While we conduct extensive validation checks, these questions may differ from actual questions that may be asked in pedagogical settings, different both in format or content.
Our study categorizes learners into only three broad educational backgrounds (elementary, high school, graduate), whereas real learners exist on a continuum of knowledge levels, with varying prior expertise and learning needs.
There might also be potential overlaps in learning needs amongst learners with different educational backgrounds.
While our benchmark includes a diverse set of ``Why'' questions, our human evaluation studies are conducted on a subset of the dataset, as carefully conducted human experiments are very expensive.

\section*{Ethics Statement}

The benchmark we introduce, \dataset, will be publicly released along with all model and human curated explanations.
All user studies were conducted by participants from the U.S. 
We designed the task to compensate annotators above minimum wage (\$16.5/hour) and conducted extensive qualification rounds before task participation. 
Annotators who completed these qualification tasks received additional compensation to account for the time required to familiarize themselves with task instructions. 
We also maintained direct communication with participants to address queries and concerns. Additionally, we provided performance-based bonuses to annotators who flagged errors or consistently provided high-quality annotations.
AI Assistants (Copilot and ChatGPT) are used as assistants in coding tasks.

\bibliography{acl_latex}

\appendix

\section{Related Work}


Previous research has explored the novelty of LLM-generated explanations (for various types of question-answering tasks), assessing how much additional information an explanation provides beyond what is already contained in the question \cite{chen-etal-2023-rev, joshi-etal-2023-machine}.
Recent advancements in LLMs integrated them into educational applications such as conversational tutoring systems (CTSs)~\cite{schmucker2024ruffle}, intelligent tutoring systems (ITSs)~\cite{stamper2024enhancing}, and other AI-driven tutoring frameworks tailored especially for younger learners~\cite{strielkowski2024ai}.
Learners spend more time with these systems, integrating them into daily life, sharing even their deep emotions and daily experiences ~\cite{seo2024chacha}.
This deep bond can make these systems the first point of reference for learners, especially younger ones, whose questions increase in the presence of an easily accessible responder~\cite{tizard2008young}.
Meanwhile, teachers and parents may struggle with some of these questions~\cite{telford2021but}, or learners may find their explanations unsatisfactory~\cite{corriveau2014does}. \citet{corriveau2014does} showed that children as young as 3 to 5 could ask complex questions about physics, biology, and social science, detect circular explanations, and reject them.
They establish causal relationships~\cite{kurkul2018question} and enhance their theory-building abilities~\cite{callanan1992preschoolers,chouinard2007children} by exploring the world through `\textit{why}' and `\textit{how}' questions.
While teachers leverage LLMs to explore diverse teaching strategies~\cite{feldhus2024towards}, accommodate different learning styles~\cite{kolb2007kolb}, and tailor instruction based on insights from conversation logs~\cite{kim2024designing}, they also express concerns that learners may encounter concepts misaligned with educational goals~\cite{kim2024designing}.

Learners, in turn, often find LLM responses unsuitable due to context ignorance~\cite{lovato2019hey} or excessive length~\cite{lee2023dapie,bertrand2023selective}. For example, when asked, ``Why do polar bears have white fur?'' an AI might respond:
``Polar bears have white fur to blend into their environment. Their coat is so well camouflaged in Arctic settings that it can sometimes pass as a snowdrift. Interestingly, their fur contains no white pigment; rather, a polar bear’s skin is black, and its hairs are hollow.''~\cite{lee2023dapie}

To address these challenges, researchers have explored interactive dialogues~\cite{lee2023dapie}, tutoring mechanisms~\cite{roscoe2008tutor} (such as learning-by-teaching~\cite{schmucker2024ruffle}), and co-creative storytelling~\cite{zhang2024mathemyths,ye2024storypark} to foster effective pedagogical interactions. ~\citet{lee2023dapie} proposed 23 guidelines for tailoring explanations for children, including the use of examples, personifications, and prompts for critical thinking. Recently, ~\citet{kim2024evallm} envisioned a human-AI interaction framework where user-specified criteria guide LLM outputs.

Our research bridges this gap by identifying mismatches between desired and generated explanations for the groups of interest, investigating their causes, and proposing solutions.
This is crucial as LLMs become more integrated into daily life, particularly for children who interact with AI through text-based interfaces and voice assistants like AI agents in home~\cite{lovato2019hey}.
Improving explanation-tailoring methods could enable a dedicated for children mode in such devices, ensuring more age-appropriate and effective responses.

\section{\dataset Dataset}
\label{sec:apx:dataset}

\subsection{Generation questions from \gptfour}
\begin{table*}[htbp]
    \small
    \centering
    \resizebox{\textwidth}{!}{
    \begin{tabular}{ccp{0.22\textwidth} @{\hspace{20pt}} >{\itshape\raggedright\arraybackslash}l}
        \hline
        \textbf{ID} 
        & \textbf{Domain} & \textbf{Discipline} 
        & \textbf{Question Text} \\
        \hline
1 & STEM & Physics & Why does thunder make a noise? \\
2 & STEM & Biology & Why do flies like poop? \\
3 & Non-STEM & Linguistics & Why are there so many languages in the world? \\
4 & Non-STEM & Earth science & Why are there waves in the ocean? \\
5 & STEM & Biology & Why do we need sleep? \\
6 & STEM & Biology & Why do leaves change color in fall? \\
7 & Non-STEM & Psychology & Why do we dream? \\
8 & STEM & Biology & Why is human birth more difficult than for other animals? \\
9 & STEM & Biology & Why do our nails grow? \\
10 & Non-STEM & Economics & Why are coins round? \\
11 & STEM & Physics & Why is glass transparent since it is made from the same thing as sand? \\
12 & Non-STEM & Psychology & Why do people bite their nails? \\
13 & Non-STEM & History & Why is number 13 considered unlucky? \\
14 & STEM & Biology & Why are eggs egg-shaped? \\
15 & STEM & Biology & Why did the dodo die out? \\
16 & STEM & Engineering and technology & Why are manhole covers round? \\
17 & STEM & Physics & Why does it echo if we yell in a cave but not a regular room? \\
18 & STEM & Biology & Why do some animals live longer than others? \\
19 & STEM & Biology & Why are polar bears white? \\
20 & Non-STEM & Geography & Why are there so many countries in the world? \\
21 & STEM & Biology & Why do we itch? \\
22 & Non-STEM & Sociology & Why do fashions change? \\
23 & Non-STEM & Sociology & Why do people get divorced? \\
24 & STEM & Physics & Why is the sky blue? \\
25 & STEM & Engineering and technology & Why do fridges hum? \\
26 & STEM & Psychology & Why do people do drugs? \\
27 & STEM & Physics & Why are snowflakes hexagonal? \\
28 & Non-STEM & Anthropology & Why do we shake our heads for "no"? \\
29 & STEM & Architecture and design & Why are most clocks round? \\
30 & STEM & Astronomy & Why does Saturn have rings? \\
31 & STEM & Earth science & Why did the dinosaurs die out? \\
32 & STEM & Biology & Why do we hiccup? \\
33 & STEM & Biology & Why do lions roar? \\
34 & Non-STEM & Sociology & Why do some people want tattoos? \\
35 & STEM & Biology & Why are dogs loyal? \\
36 & Non-STEM & Psychology & Why does tickling make us laugh?  \\
37 & Non-STEM & Psychology & Why do some people bully? \\
38 & STEM & Biology & Why do our noses run when we eat spicy food? \\
39 & STEM & Biology & Why are lemons sour? \\
40 & STEM & Biology & Why are we awake during the day and sleepy at night? \\
41 & Non-STEM & Psychology & Why are women often more emotional than men? \\
42 & STEM & Physics & Why is water transparent? \\
43 & STEM & Biology & Why are honeycombs hexagonal? \\
44 & Non-STEM & Architecture and design & Why are jeans blue? \\
45 & STEM & Biology & Why are flowers colorful? \\
46 & STEM & Engineering and technology & Why are flags rectangular? \\
47 & Non-STEM & Psychology & Why do people fall in love? \\
48 & Non-STEM & Psychology & Why do people lie about small things? \\
49 & Non-STEM & Psychology & Why do we look around when we hear a noise? \\
50 & STEM & Medicine and health & Why do people die? \\
        \hline
    \end{tabular}
    }
    \caption{A set of 50 seed questions categorized by domain (STEM or Non-STEM) and academic discipline. Among these, 32 questions belong to STEM disciplines, while 18 fall under Non-STEM disciplines. 
    }
    \label{tab:seed questions}
\end{table*}

\begin{table*}[h!]
    \centering
    \begin{tabular}{p{0.95\textwidth}}
        \toprule
        \textbf{Model:} \gptfour-0613, \textasciitilde 1.8 trillion parameters \\ \midrule
        \textbf{max\_tokens}: 4096 \\
        \textbf{temperature}: 1.0 \\
        \midrule
        \textbf{Prompt:} Generate 100 non-STEM "why" questions. \\
For example, STEM "why" questions can be questions about Physics, Chemistry, Biology and Neuroscience. Non-STEM "why" questions can be about Humanities, Liberal Arts, Psychology, Law, Sociology or Socio-Cultural topics to name a few domains.\\
Some examples of non-STEM questions are - \\
1. Why do people bite their nails? \\
2. Why are there so many languages in the world?\\
        \bottomrule
    \end{tabular}
    \caption{Model Configuration and Prompt Used for Overgenerating ``Why'' Questions. The domain (STEM and Non-STEM) and the in-context examples are changed multiple times to get diverse ``Why'' questions.}
    \label{tab:question_generation_config_prompt}
\end{table*}


\Cref{tab:seed questions} shows the 50 ``Why?'' questions (from \citet{sulik2023explsinthewild}) that were used as seed examples to guide our question generation for \dataset.
Refer to \Cref{tab:question_generation_config_prompt} for the configuration and prompt used for generating `Why' questions.

\subsection{Filters after question generation}
\label{sec:apx:question_filter}
After generating the questions, we manually reviewed all 30,671 instances to filter out invalid or toxic entries, including deduplicating ``similar". questions The filtering process took $\sim 12$ hours.
This human annotation process, rather than a rule-based system, ensured that only high-quality, non-toxic and non-hallucinated questions were retained. 
Additionally, we further filter questions if they are \textit{too niche} for a given domain.
This is evaluated using an ``answerability'' task performed by participants recruited from Amazon Mechanical Turk. 
Annotator's were posed the following question -- \textit{Given ``Why'' questions about world phenomena, you have to judge whether you can answer the question on your own with some help from external resources}.
We determined that questions which cannot be answered by lay annotators without any external resources would be too niche, and thus not suitable to be grade-tailored. 
Each question is annotated by three annotators. 
We only keep questions where all annotators agree that they would be able to answer the question.
As detailed in \Cref{tab:filtered_out_questions}, these ``Why'' questions were excluded from the final dataset because annotators determined that they were too niche.

\begin{table}[t]
\centering
\begingroup
\fontsize{8}{10}\selectfont
\begin{tabular}{l}
\toprule
\textbf{Question} \\ 
\midrule
Why is P always less than or equal to NP in complexity theory?\\ 
Why does Black-Scholes model matter to finance?\\ 
Why is quantum entanglement paradoxical?\\ 
Why was Euclid's fifth postulate so controversial?\\ 
Why do carbon atoms form four bonds in organic chemistry?\\ 
Why do DNA strands run from the 3' to 5' direction?\\
\bottomrule
\end{tabular}
\endgroup
\caption{Examples of niche questions filtered out based on annotators' answerability judgments.}
\label{tab:filtered_out_questions}
\end{table}


\subsection{Distribution of different academic disciplines in \dataset}

\begin{figure*}[ht]
    \centering
    \includegraphics[width=\textwidth]{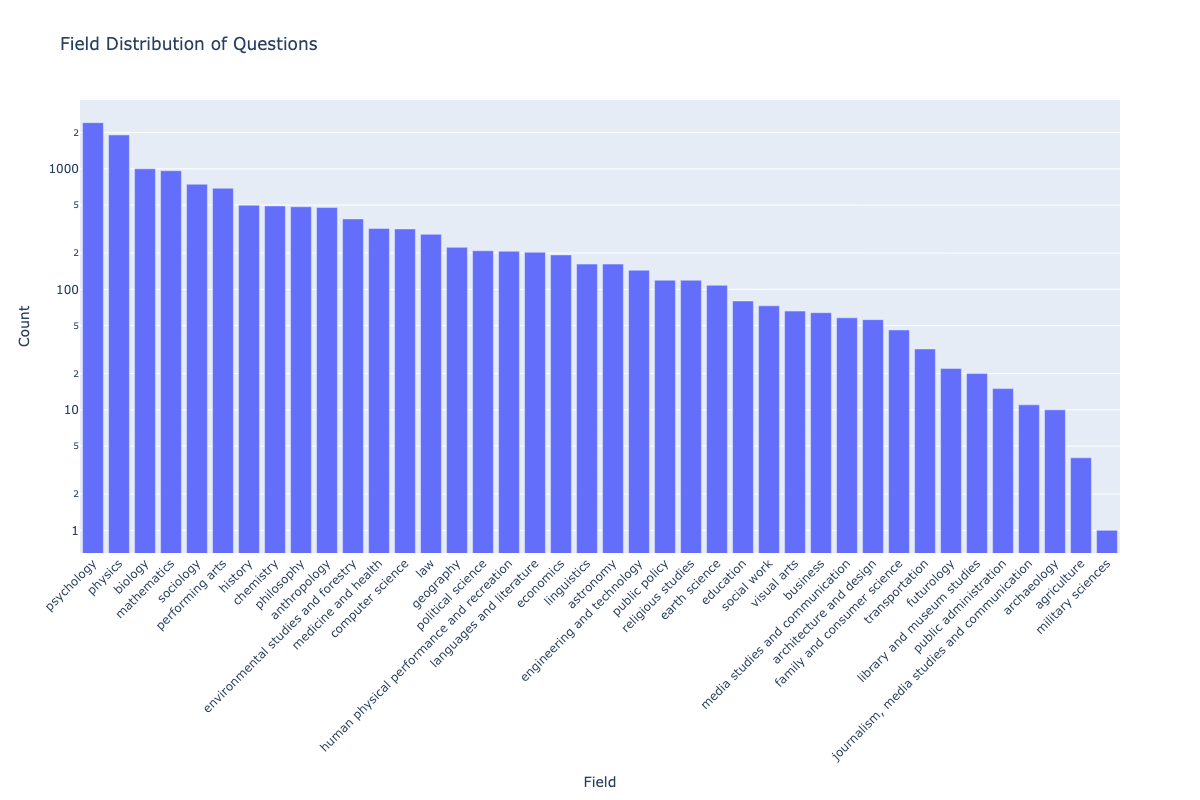}
    \caption{The distribution of question fields.} 
    \label{fig:field_distribution_full}
\end{figure*}

\begin{table*}[h!]
    \centering
    \begin{tabular}{p{0.95\linewidth}}
        \toprule
        \textbf{model:} \textbf{\llama-4bit} and \textbf{\llamaseventy-4bit} \\ \midrule
        \textbf{max\_tokens}: 50 \\
        \textbf{temperature}: 0.2 \\
        \textbf{seed:} 0 \\
        \textbf{GPU}: Apple M4 Max \\
        \textbf{Inferring time}: 8 hours \\ \midrule
        \textbf{Prompt:} You are a helpful assistant that categorizes questions into their relevant domains. Please classify each question into the most related domain in the following domains exactly as written (in lowercase): [discipline list]. Respond with only the domain name. \\
        \midrule
        \textbf{Fallback Prompt:} You are a helpful assistant that categorizes questions into their relevant domains. Please classify each question into the most related domain in the following domains exactly as written (in lowercase): [discipline list]. Respond with only the domain name. If the domain is not in the list, please provide your best guess in one word or a short phrase (provide the field only without extra words). \\
        \bottomrule
    \end{tabular}
    \caption{Model Configuration and Prompting Details for Discipline Classification. The Fallback Prompt will be used on \llamaseventy if we prompted both \llama and \llamaseventy 20 times each and none responds with a class in the list.}
    \label{tab:discipline_model_config}
\end{table*}

For classifying the question fields, we used a classification task with a 
reference list of fields derived from the ``Outline of Academic Disciplines'' (using its second subheader)\footnote{Subheadings outlined in \url{https://en.wikipedia.org/wiki/Outline_of_academic_disciplines}}. We run this task with \llama and \llamaseventy with the prompt in \Cref{tab:discipline_model_config}; 
We run both models 20 times each; if Llama did not assign a question to one of the provided fields, we prompted it to return the most suitable field it could find, and manually classified it to the closest field.

~\Cref{fig:field_distribution_full} shows all the fields classified by \llamaseventy.
We performed a sanity check of the field classification by performing a manual check of $50$ questions. 

\section{Generating Explanations for \dataset}
\label{sec:apx:expl_generation}

\subsection{Prompts used for explanation generation}

\begin{table*}[ht]
    \footnotesize
    \centering
    \resizebox{0.98\linewidth}{!}{
    \begin{tabular}{p{0.9\textwidth}}
        \toprule
        \textbf{Dataset Collection Prompt:}\\
        Generate 100 [STEM/Non-STEM] "why" questions.\\
        For example, STEM "why" questions can be questions about Mathematics, Deductive Reasoning, Logical Reasoning, Physics, Chemistry, Biology and Neuroscience. Non-STEM "why" questions can be about Humanities, Liberal Arts, Psychology, Law, Sociology or Socio-Cultural topics to name a few domains. \\[2mm]
        \midrule
        \midrule
        
        \textbf{\phd Prompt:} \\
        You will be asked a "Why" question. You are an expert in the domain of the why question you are asked. The user asking you the question also has a PhD in the domain of the question they asked. 

        Your job as an expert is to provide a concise explanation to the PhD holder. Make sure that the explanation is useful to the user - they will use it to validate and cross check important information. They may also use the explanation to teach that topic to a class.
        
        Just provide the explanation as is - do not add any additional text like greetings or ornamental words.\\
        \\

        \textbf{\hs Prompt:} \\
        You will be asked a "Why" question. You are an expert in the domain of the why question you are asked. The user asking you the question is someone who holds a basic american high school education. You can assume they are a "layperson" to the domain of the question asked.
        
        Your job as an expert is to provide a concise explanation to the user. They asked you the question as they were curious about the topic, so make sure that the explanation is useful to them.
        
        Just provide the explanation as is - do not add any additional text like greetings or ornamental words.\\
        \\

        \textbf{\es Prompt:} \\
        You will be asked a "Why" question. You are an expert in the domain of the why question you are asked. The user asking you the question is someone who holds a basic american elementary school education. 

        Your job as an expert is to provide a concise explanation to the user.
        
        Just provide the explanation as is - do not add any additional text like greetings or ornamental words.\\
        \\

        \textbf{\default Prompt:} \\
        You will be asked a "Why" question. You are an expert in the domain of the why question you are asked. 

        Your job as an expert is to provide a concise explanation to the user. 
        
        Just provide the explanation as is - do not add any additional text like greetings or ornamental words.\\
        \bottomrule
    \end{tabular}}
    \caption{Prompts used for Dataset Collection and Explanation Generation}
    \label{tab:prompts_used}
\end{table*}

\Cref{tab:prompts_used} presents the complete set of prompts used for dataset collection and explanation generation.
\gptfour has a proprietary license.
While all our analysis in this paper are based on \gptfour, we also generate explanations from \llama to conduct automatic evaluation \Cref{sec:apx:auto_score_dist}.

\subsection{Model details}
\label{sec:apx:model_details}

\Cref{tab:explanation_generation_config} shows the configurations of \gptfour and \llama used during explanation generation.

\begin{table*}[h]
    \centering
    \resizebox{\linewidth}{!}{
    \begin{tabular}{cllll}
        \toprule
        \textbf{Config} & \textbf{\gptfour Assignment} & \textbf{Llama Assignment} & \textbf{Qwen Assignment} & \textbf{Deepseek R1 Distilled Assignment}\\
        \midrule
        \multirow{2}{*}{model} 
        & \textbf{gpt-4-0613} & \textbf{\llama} & \textbf{\qwen} & \textbf{\deepseek} \\
        & Number of parameters: \textasciitilde 1.8 trillion
        & Number of parameters: 3 billion
        & Number of parameters: 14 billion
        & Number of parameters: 8 billion\\
        \midrule
        \midrule
        max\_tokens & 4096 & 4096 & 4096 & 4096\\
        temperature & 1 & 0.1 & 0.1 & 0.1\\
        seed & random & 0 & 0 & 0\\
        GPU & N/A, openai api call & A100 & A100 & A100\\
        Inferring time & N/A & 2 hours & 4 hours & 4 hours\\
        \bottomrule
    \end{tabular}
    }
    \caption{Model Configurations for Explanation Generation}
    \label{tab:explanation_generation_config}
\end{table*}

\section{Human Experiments}
\label{sec:apx:human_exp}
Participants were recruited from Prolific\footnote{\url{https://www.prolific.com/}}, who consented to our study.
Participants were given an option to exit the study at any point.
For Perceived Background Match evaluation in \Cref{sec:perceived:human_eval}, participants were screened based on location (United States), education level (high school or higher), and active participation to ensure high-quality responses.
For informativeness simulation, participants were also recruited through Prolific and screened based on location, educational background, discipline of study (for graduate-level participants), and active participation on the platform. 
To ensure high-quality responses, all participants received detailed task instructions and were provided with examples to clarify expectations.
Additionally, participants submitted natural language justifications for their selections, allowing further insight into their reasoning. 
Each participant annotated five questions, and each question received five independent annotations for each educational background.
There were a total of 811 unique participants in all our studies.

\subsection{Annotator Filtering Criteria} 
We applied task-specific filtering criteria during participant recruitment. Table~\ref{tab:annotator_filtering} details the screener settings for each group—Elementary School, High School, Physics Graduates, and Psychology Graduates. All annotators were required to reside in the US and have English as their primary language.

\begin{table*}[t]
    \centering
    \small
    \begin{tabularx}{\textwidth}{%
        >{\raggedright\arraybackslash}p{0.17\textwidth} | 
        >{\centering\arraybackslash}p{0.17\textwidth} 
        >{\centering\arraybackslash}p{0.18\textwidth} 
        >{\centering\arraybackslash}p{0.18\textwidth} 
        >{\centering\arraybackslash}p{0.18\textwidth} }
        \toprule
        \textbf{Config} & \textbf{Elementary School} & \textbf{High School} & \textbf{Physics Graduates} & \textbf{Psychology Graduates} \\
                        & \textbf{Screener}          & \textbf{Screener}    & \textbf{Screener}          & \textbf{Screener} \\
        \midrule
        Current Country of Residence & 
        United States & United States & United States & United States
        \\
        Primary Language & English & English & English & English\\
        Age & n/a & 18-24 & n/a & n/a\\
        Approval Rate & 98-100 & 98-100 & n/a & n/a \\
        Number of previous submissions & 1000-10000 & 1000-10000 & n/a & n/a\\
        Highest education level completed & No formal qualifications & High school diploma/A-levels, Technical/community college & Graduate degree (MA/MSc/MPhil/other), Doctorate degree (PhD/other) & Graduate degree (MA/MSc/MPhil/other), Doctorate degree (PhD/other) \\
        Degree subject & n/a & n/a & Natural Sciences & Psychology \\
        \bottomrule
    \end{tabularx}
    \caption{Prolific annotator filtering}
    \label{tab:annotator_filtering}
\end{table*}

\subsection{Annotators Demographic Distribution}
\begin{figure*}[t]
    \centering
    \includegraphics[width=\textwidth]{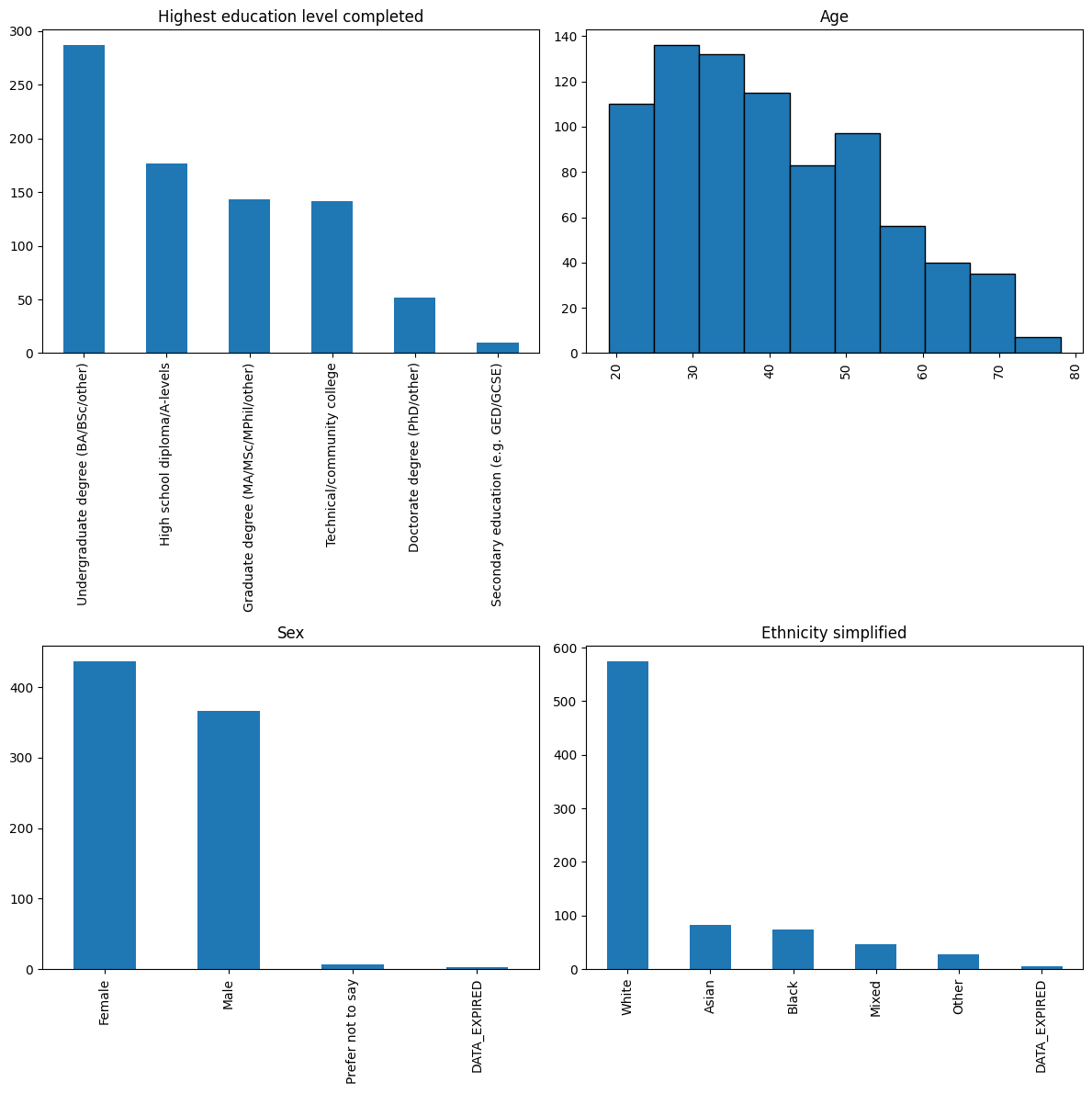}
    \caption{Annotators Demographic Distribution}
    \label{fig:prolific_demographic}
\end{figure*}

We collected demographic information, including highest education level, age, sex, and ethnicity, to ensure a representative sample. All annotators were required to have a country of residence in the United States and a primary language of English. \Cref{fig:prolific_demographic} presents the overall distribution of these demographics, confirming the diversity of our 811 unique annotators and enhancing the generalizability of our findings.

\begin{figure*}[t]
    \centering
    \begin{subfigure}{0.48\textwidth}
        \includegraphics[width=\textwidth]{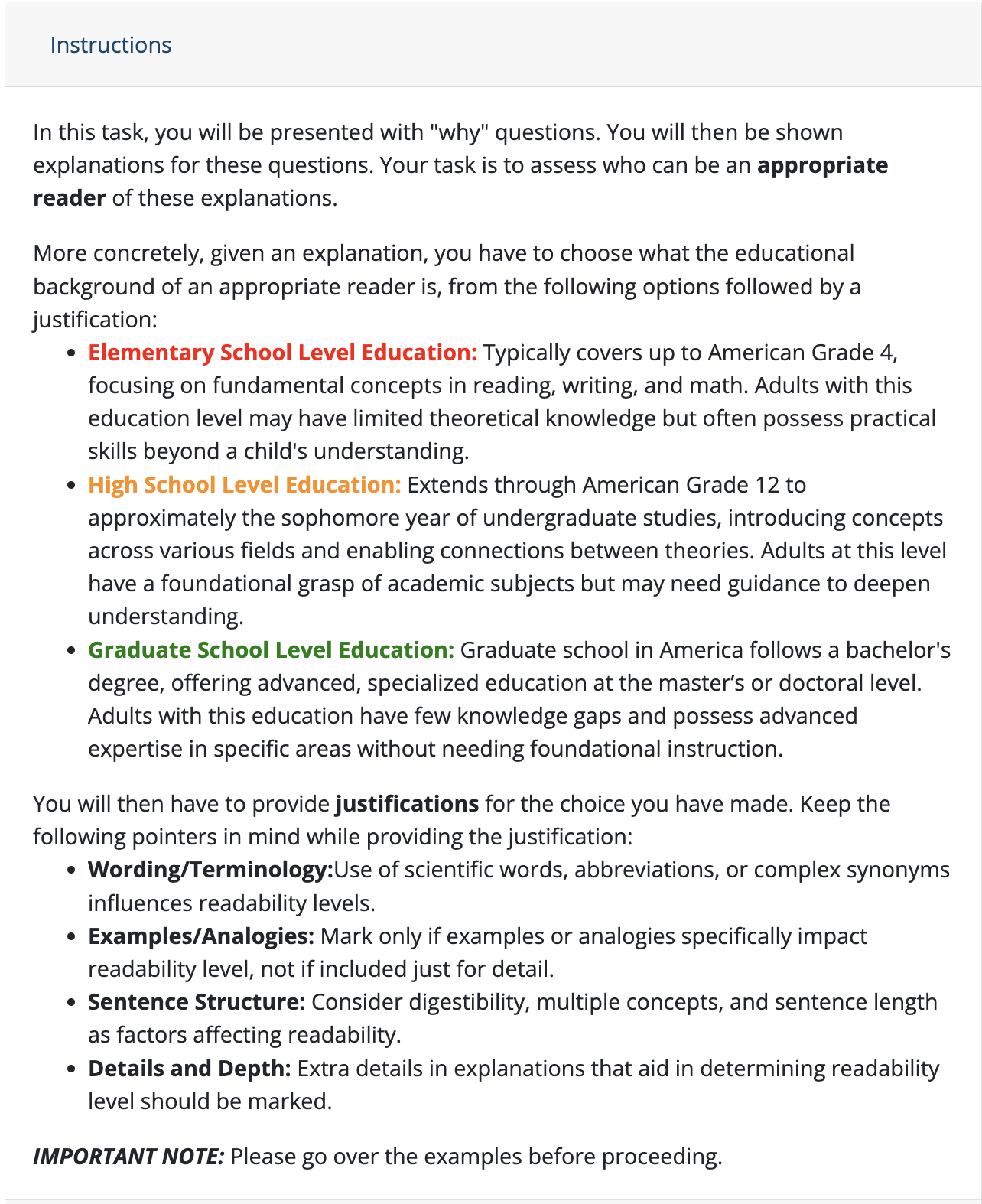}
        \caption{Instruction for perceived background match user study: We asked annotators to assess the appropriate reader of explanations.}
    \end{subfigure}
    \hfill
    \begin{subfigure}{0.48\textwidth}
        \includegraphics[width=\textwidth]{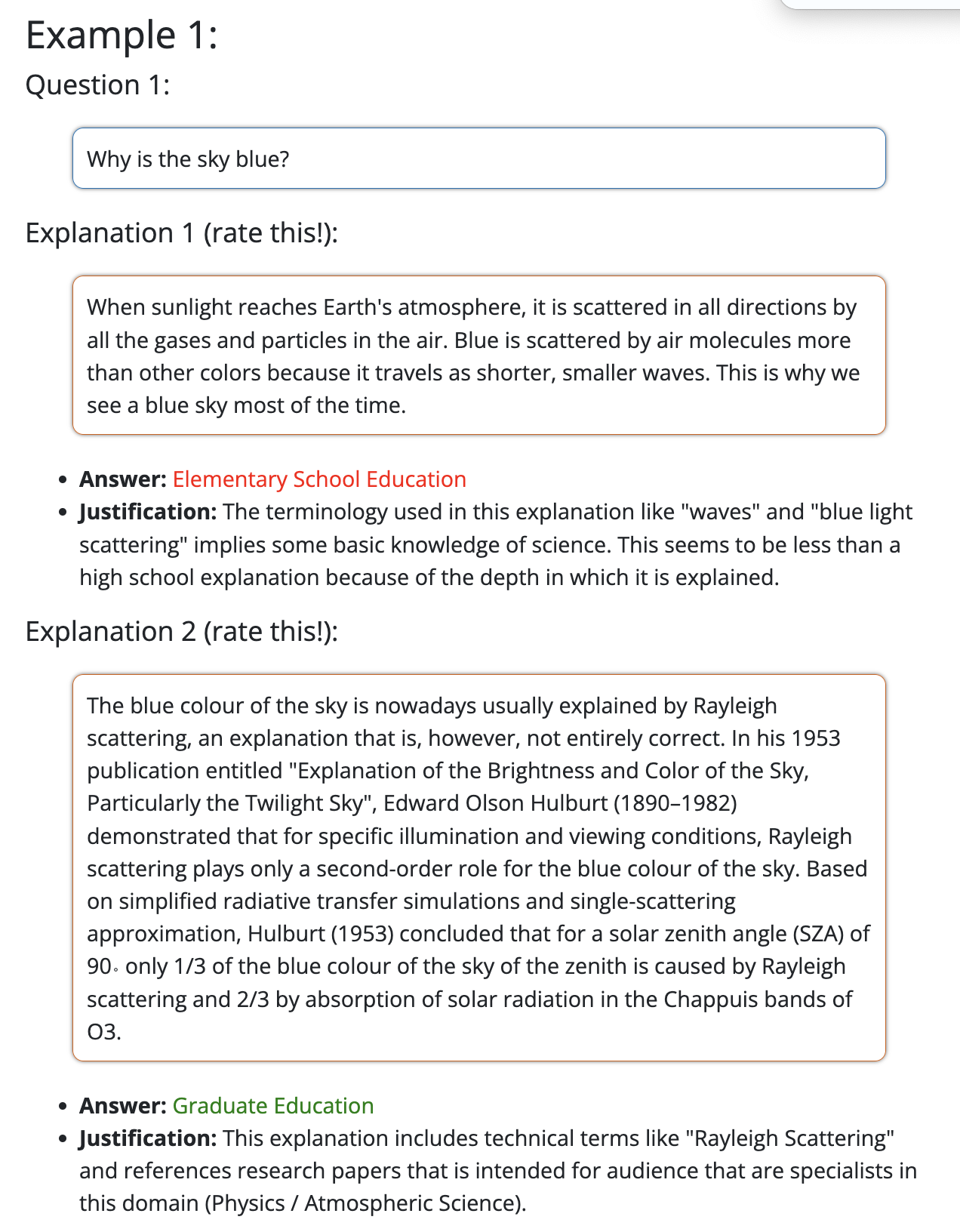}
        \caption{Example for perceived background match user study: We selected 2 questions and 5 explanations to clarify the task details.}
    \end{subfigure}

    \begin{subfigure}{0.48\textwidth}  
        \centering
        \includegraphics[width=\textwidth]{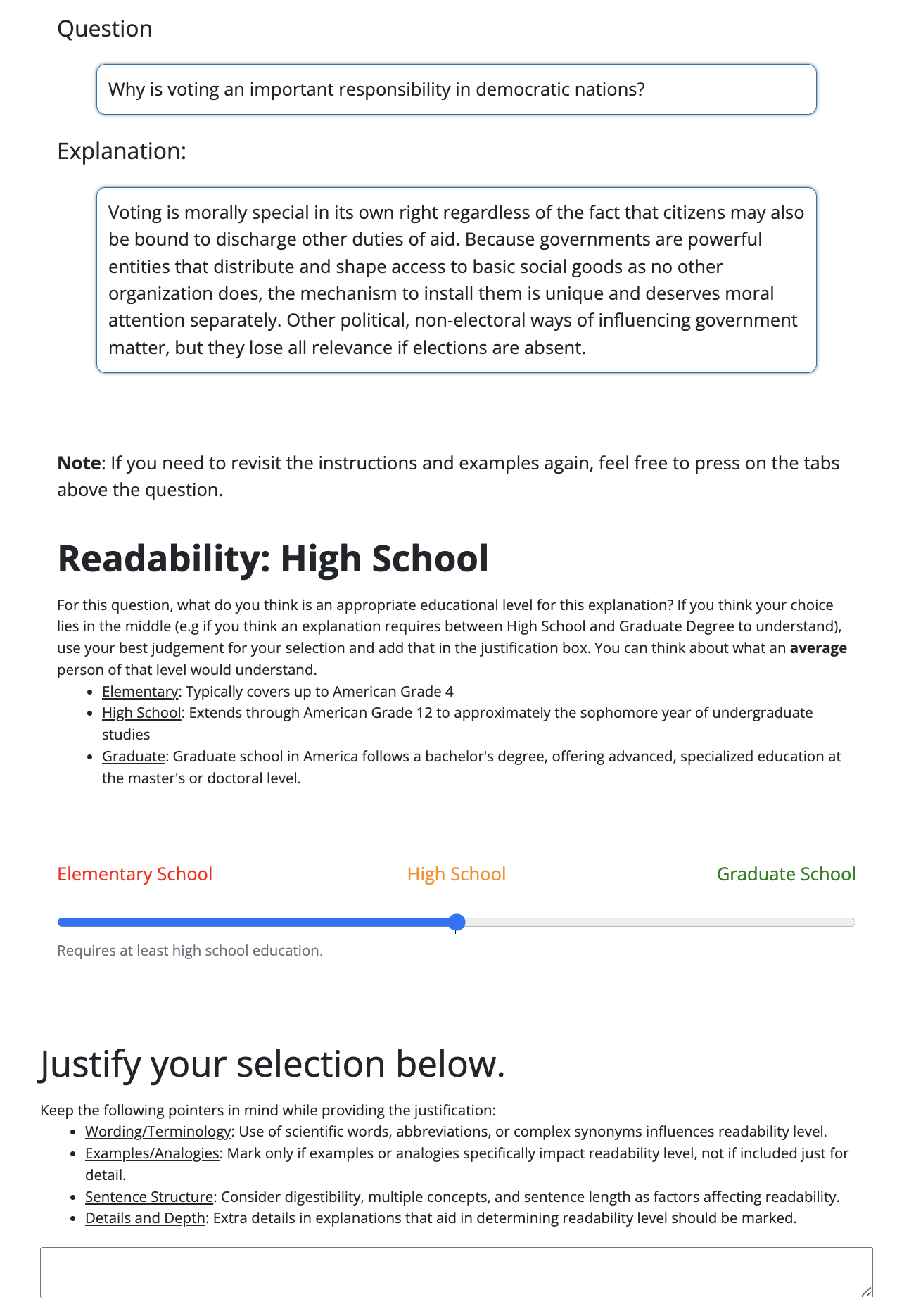}
        \caption{Question for perceived background match user study: Annotators select the readability level using a scrollbar and justify their choice based on factors such as wording, examples, sentence structure, and depth of detail.}
    \end{subfigure}

    \caption{Overview of the perceived background match user study.}
    \label{fig:perceived_readability_form}
\end{figure*}

\begin{figure*}[t]
    \centering
    \begin{subfigure}{0.43\textwidth}
        \includegraphics[width=\textwidth]{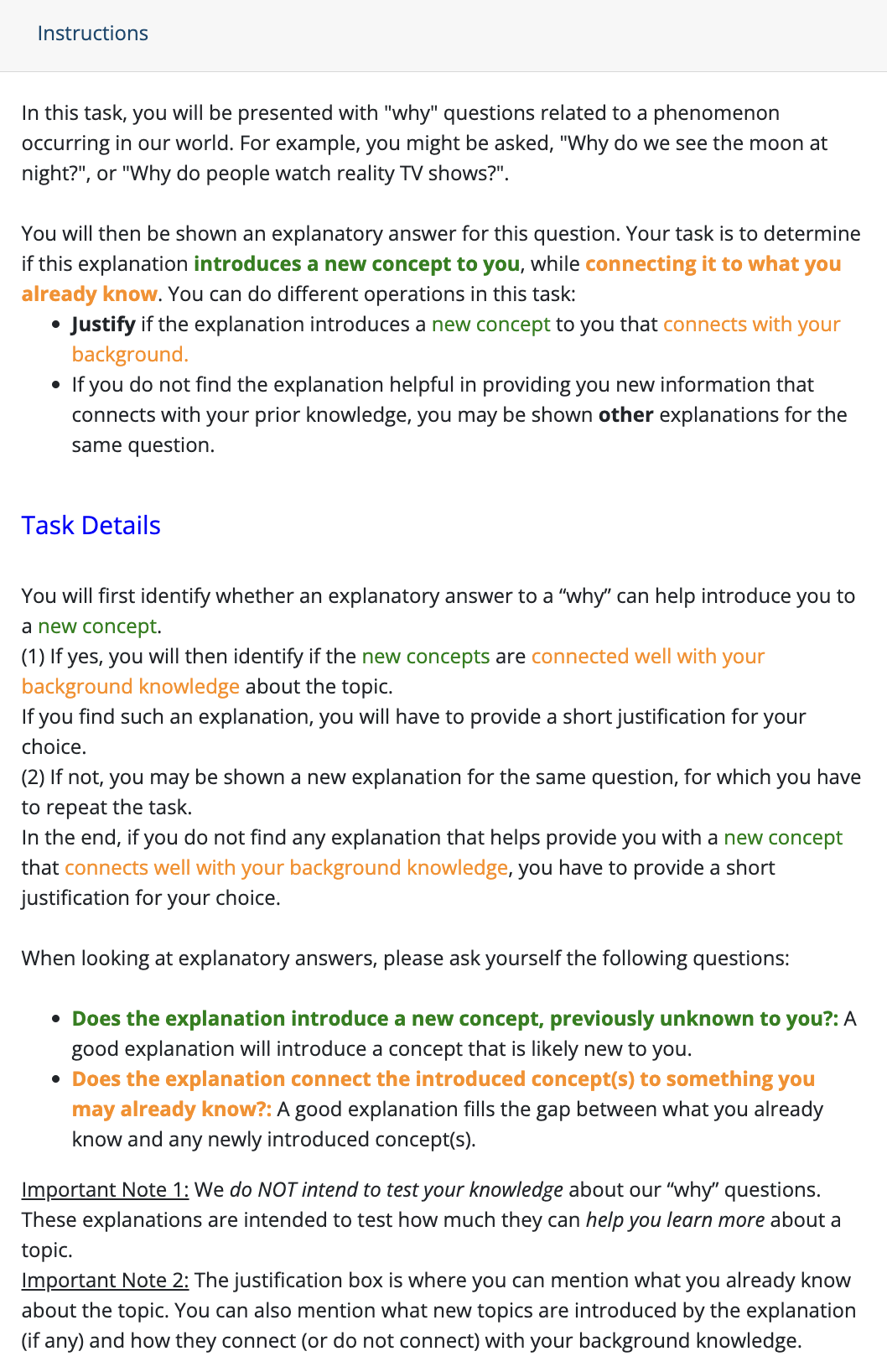}
        \caption{Instruction for explanation informativeness user study: We asked annotators to determine if the explanation introduces a new concept and connects well with their background.}
    \end{subfigure}
    \hfill
    \begin{subfigure}{0.46\textwidth}
        \includegraphics[width=\textwidth]{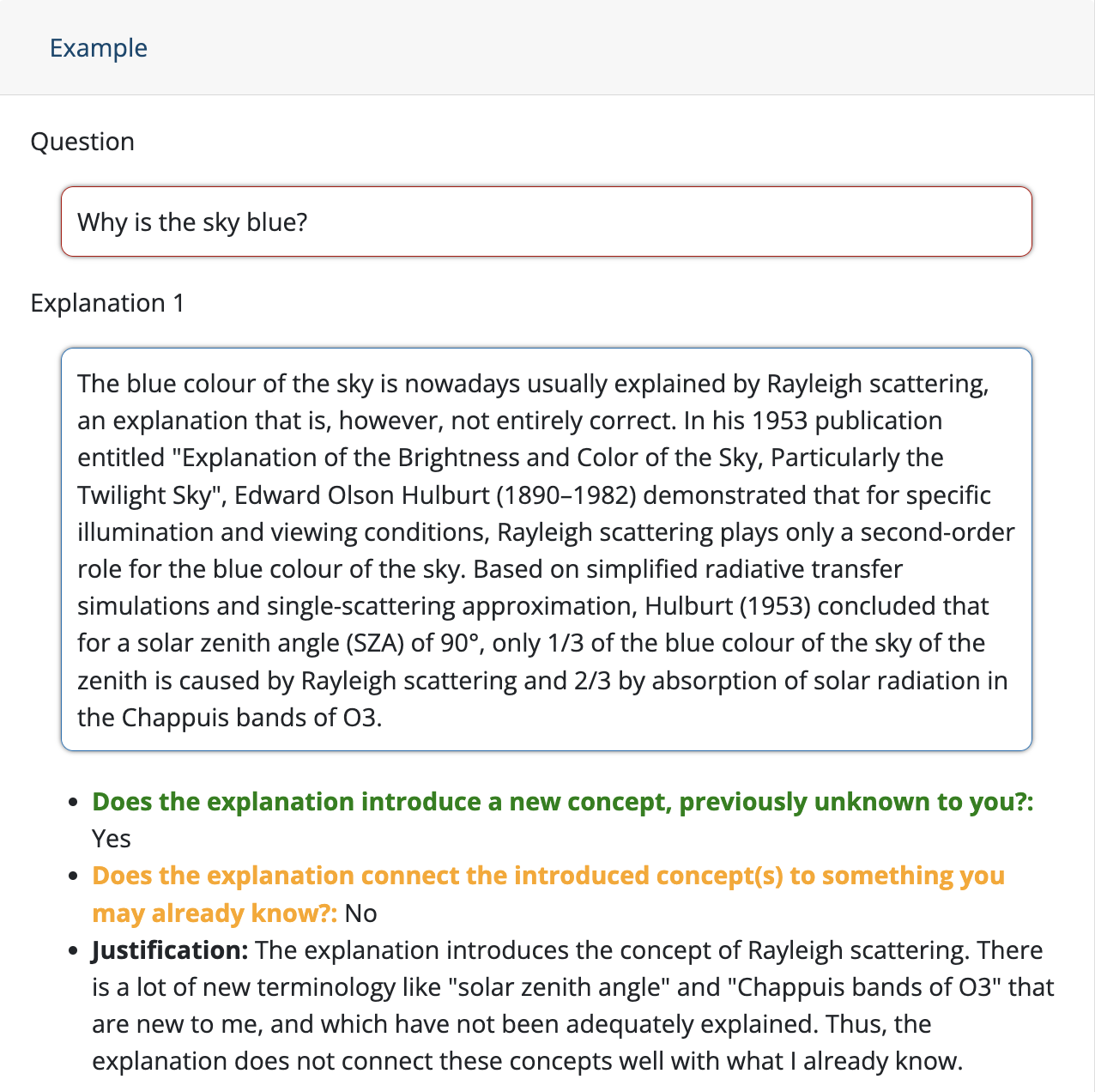}
        \caption{Example for explanation informativeness user study: We selected 3 explanations to clarify the task details.}
    \end{subfigure}

    \begin{subfigure}{0.40\textwidth}  
        \centering
        \includegraphics[width=\textwidth]{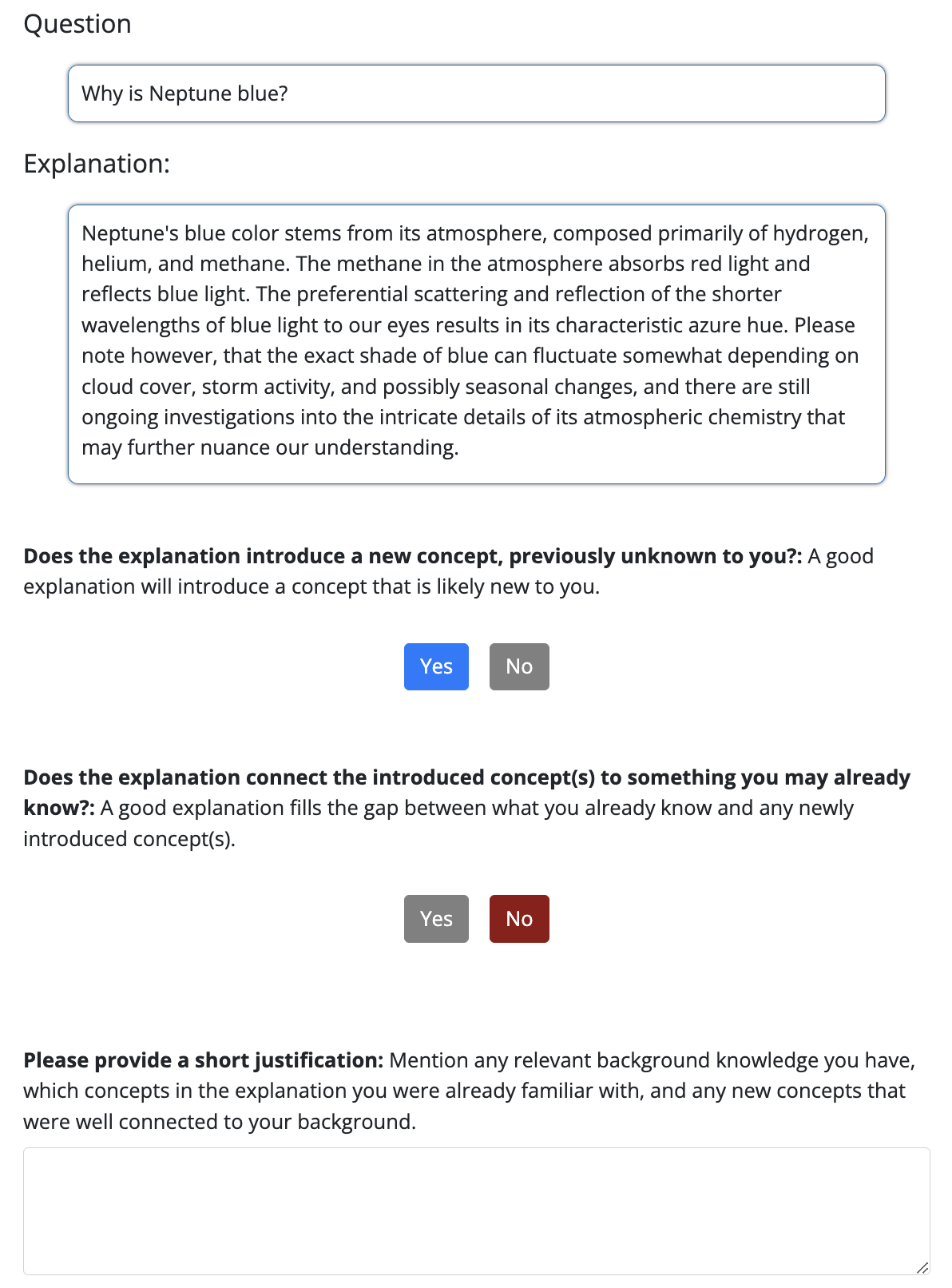}
        \caption{Question for explanation informativeness user study: Annotators select whether the explanation introduces new concepts and whether those concepts are connected to their prior knowledge. They provide justification based on their background understanding.}
    \end{subfigure}

    \caption{Overview of the explanation informativeness user study.}
    \label{fig:new_info_form}
\end{figure*}

\subsection{Overview of the User studies}

\Cref{fig:perceived_readability_form} and \Cref{fig:new_info_form} provide an overview of the two human experiments, including instructions, examples, and the evaluation interface.

\subsection{Example annotations for perceived background match user study}
\label{sec:apx:anno2.1}

\Cref{tab:2.1anno_example} presents the annotations for the perceived background match user study, with different annotators' justifications separated by semicolons.
It shows examples where annotators mark the explanation's perceived educational background based on their evaluation of the language and detail, and provided their justifications of choices. 
Note that annotators weren't aware of the intended readability level.

\Cref{tab:2.2anno_example} displays the annotations for the explanation informativeness user study, where the explanation column shows the final explanation shown to participants.
In the last two rows of \Cref{tab:2.2anno_example}, annotators entered the ``need more complex explanation'' path despite the explanation being at a graduate level, resulting in no available explanations and both the informativeness and the match being marked as ``no.''

\begin{table*}[]
\scriptsize
    \centering
    \resizebox{\linewidth}{!}{
    \begin{tabular}{p{0.11\textwidth}p{0.32\textwidth}p{0.07\textwidth}p{0.08\textwidth}p{0.32\textwidth}}
    \toprule
    \textbf{Question} & \textbf{Explanation} & \textbf{Intended Educational Background} & \textbf{Perceived Educational Background (Majority)} & \textbf{Justification}\\ 
    \midrule
    \midrule
    Why does chocolate melt in the mouth? & Chocolate melts in the mouth because it's made with cocoa butter, a type of fat that melts at or just below human body temperature. So when you put chocolate in your mouth, the heat from your body is enough to melt the cocoa butter and that's why the chocolate softens and dissolves. & \esshort & \esshort & Sentences are very easy to understand as well as concepts.; Uses basic words and simple sentence structure. Doesn't go overboard with detail; It uses simple sentence structure and wording and add extra detail to make it easily understood.\\
    Why does the sun rises in the east and sets in the west? & The sun rises in the east and sets in the west because of the Earth's rotation. The Earth rotates on its axis, an imaginary line that runs from the North Pole to the South Pole. It rotates in a counterclockwise or eastward direction. Because of this rotation, it appears to us like the Sun is moving across the sky from east to west when in reality, we're the ones moving. This phenomenon is called apparent solar motion, and it's the reason for the sun's apparent path in the sky. & \hsshort & \esshort & The wording and details suggests elementary school.; Wording, details and depth; The examples given are straight to the point and do not go into too much depth.  The details of the reasoning are not too long or difficult to follow. The sentence structure is short. The use of words like axis, North Pole, South Pole, imaginary, etc. is not of an high school level, but does show that they have knowledge of science.\\
    
    Why are people's blood types different? & Blood types are different due to the presence or absence of certain molecules, called antigens, on the surface of red blood cells. These antigens are essentially identifiers which your immune system use to recognize its own cells. The different blood types, A, B, AB, and O, refer to these specific antigens. Your blood type is determined by the genes you inherit from your parents. & \esshort & \hsshort & The wording was simple as well as the sentence structure and details; This explanation uses more advanced terminology like molecules and antigens. But it does not go into too much detail about how they are different.\\
    Why is there less gravity on the moon? & Gravity at a specific location is determined by the mass of the object and the distance from its center. The Moon is much smaller than Earth and has less mass. This reduces the gravitational pull that it has on objects on its surface. So, when you're on the Moon, you weigh less than you do on Earth because the Moon's gravitational pull is weaker due to its smaller mass. & \hsshort & \hsshort & This explanation uses terms that would be understandable to a high school level reader but too advanced for elementary reading. The comparison of the moon's density to earth's and the explanation of how gravity operates less on mass is a good detailed response for a high school level reader.; The reading level appears to be that of high school level students. It doesn't go deeply into mathematical details beyond a high school level. It doesn't provide advanced example of why you weigh less on the moon compared to earth.\\
    Why do our bodies produce sweat when we're hot? & Sweating is the body's essential cooling mechanism. When your core body temperature rises, your hypothalamus, which is the body's thermostat, signals your sweat glands to start producing sweat. It's predominantly composed of water, but also includes minor amounts of various salts and urea. As the sweat evaporates from your skin, it helps to dissipate the heat and cool your body down. This process, known as evaporative cooling, maintains the body within a narrow, optimal temperature range, critical for the normal functioning of physiological processes and homeostasis. & \phdshort & \hsshort & The sentence structure is digestible, and topic is simple.  Terminology like "predominantly," "dissipate," "critical," "physiological processes and homeostatis" indicate more advanced than elementary.  However, not enough detail/sources to reach graduate level; Word usage is above elementary level, but should be fully comprehended by high school students.\\

    Why is the human brain folded? & The human brain is folded into wrinkles, also known as sulci and gyri, to increase the surface area for more neurons. More neurons mean more processing power and cognitive abilities. The folding allows our brains to fit a larger number of neurons within the confined space of our skulls. Like squeezing more electronic circuits into a smaller computer chip, the folds allow for greater complexity and speed in our thinking process. & \hsshort & \phdshort & This article introduced specialized terminology related to the brain, like gyri. The article assumed that the reader would easily understand how increased surface area would lead to increased processing using only an analogy of a computer chip, which also requires preexisting electronics knowledge.; The sentences contain terminologies that are appropriate for professional audience or specialists \\

    Why do fruits turn sweet when they ripen? & Fruits ripen due to the action of a plant hormone called ethylene. During the ripening process, ethylene accelerates the conversion of starches to sugars in the fruit, leading to an increased sweetness. Complex carbohydrates are broken down into simpler sugars; for instance, sucrose is converted into glucose and fructose. This is a part of the plant's strategy to make the fruit appealing to animals, so they will eat it and subsequently disperse the seeds in the process. Additionally, the process of ripening also often involves other changes, such as color, texture, aroma development, or an increase in nutritional content, which are conducive to seed dispersal and survival. & \phdshort & \phdshort & Detailed examples sounds like someone who may have advanced schooling in the subject.; I picked grad level, because it assumes prior knowledge of higher level science concepts.\\
    \bottomrule
    \end{tabular}
    }
    \caption{Annotations for perceived background match user study, different annotator's justifications are separated by semicolons.
    }
    \label{tab:2.1anno_example}
\end{table*}
\begin{table*}[]
\small
    \scriptsize
    \resizebox{\linewidth}{!}{
    \begin{tabular}{p{0.08\textwidth} p{0.35\textwidth}p{0.08\textwidth}p{0.04\textwidth}p{0.04\textwidth}p{0.3\textwidth}}
    \toprule
    Question & Explanation & Intended Educational Background & Informa-tive & Matched & Justification\\
    \midrule
    \midrule
    \multicolumn{6}{l}{\es Background Annotators}\\
    \midrule
    Why do we yawn? &
    We yawn as a way to wake up. When you're tired or bored, your body takes in less oxygen because your breathing slows down. Yawning helps you take in a lot of air and uses muscles which sends signals to your brain to wake it up. It also helps cool the brain. It's even contagious because it's a primitive reflex humans have adapted for social bonding and empathy. &
    \esshort & 
    Yes &
    Yes &
    I knew we yawn for empathetic reasons from some videos on psychology. And of course, I knew we yawn when we're tired or bored. I did not, however, know that we take in less oxygen from our slowed breathing when we're tired, and that is why we yawn. I also didn't know it sent signals to the brain to wake it that way.\\
    
    Why does acid rain occur? &
    Acid rain occurs due to the release of sulfur dioxide (SO2) and nitrogen oxide (NOx) into the atmosphere. These gases are primarily released by burning fossil fuels like coal, oil, and gas in power plants, cars, and factories. Once in the atmosphere, they react with water, oxygen, and other substances to form sulfuric acid and nitric acid. These acids can then be carried over long distances by wind and eventually fall as acid rain. This process can be harmful to the environment, damaging vegetation, aquatic habitats, and even buildings. &
    \hsshort &
    Yes &
    No &
    While I knew acid rain can be caused from fossil fuel burning, I did not know the types of acid it produces and how it is actually formed from mixing with water, oxygen and other substances. It reinforces the damage that it can cause to the environment that I was also aware of.\\
    
    \midrule\midrule
    \multicolumn{6}{l}{\hs Background Annotators}\\
    \midrule
    Why does DNA have a double helix structure?&
    DNA has a double helix structure because of the nature of the molecules it's made of and the physical and chemical interactions between them. It consists of two strands of nucleotides, each made up of a sugar, a phosphate, and a base. The sugar and phosphate form the backbone of the DNA, and the bases (adenine, cytosine, guanine, thymine) from each strand bond together. Adenine always pairs with thymine, and cytosine with guanine. These pairs are of almost equal shape and size, generally fitting neatly between the two spiraling backbones, maintaining a constant distance, thus ensuring the helical structure. Furthermore, the double helix provides protection to the genetic information and is more stable, which is crucial for genetic inheritance.&
    \hsshort &
    Yes &
    Yes &
    I had learned much about DNA in compulsory education but had forgotten the real reason for the double helix shape. This introduced to me the bases adenine,cytosine,guanine and thymine being in the mix. \\
    
    Why do we have day and night? &
    Day and night occur due to the Earth's rotation on its axis. The Earth spins around an imaginary line that passes through the North and South Poles, known as the Earth's axis. It completes one full rotation in approximately 24 hours. 

    During this rotation, only the half of the Earth that's facing the sun experiences day, as it's being illuminated by sunlight. Simultaneously, the other half of Earth faces away from the sun and is shrouded in darkness, and so it experiences night. 
    
    As the Earth continues rotating, the regions exposed to sunlight shift, so different parts of the world pass through day and night cycles. This constant rotation creates the 24-hour day and night cycle we experience. &
    \phdshort &
    No &
    No &
    The explanation does not introduce a completely new concept to me, as I was already familiar with the idea that Earth's rotation causes day and night. However, it reinforced my understanding by clearly describing the Earth's axis as an imaginary line passing through the North and South Poles. The explanation also effectively connected the idea of Earth's continuous rotation to the 24-hour cycle we experience. While the core concept was familiar, the phrasing and details helped refine my understanding.\\
    
    \midrule
    \midrule
    \multicolumn{6}{l}{\phd (Psychology) Background Annotators}\\
    \midrule
    Why do we sleep?&
    Sleep serves multiple purposes that are essential to the human brain and body. 

    One of the key reasons we sleep is for restoration. During sleep, the body has the chance to repair muscle tissue and synthesize proteins, restore energy, and release the growth hormone which is crucial for growth and development. 
    
    In the brain, sleep helps in clearing out waste products, some of which are potentially neurotoxic, that brain cells generate as a byproduct of their activity. This clearance is carried out by the glymphatic system which becomes especially active during sleep.
    
    Sleep also has a significant role in memory consolidation. The neural connections that form our memories are strengthened during sleep, particularly during REM sleep. During this time, the brain reorganizes and restructures memories, which may result in more creativity.
    
    Lastly, lack of sleep can cause numerous health problems, including weakened immunity, increased risk of chronic diseases, poor mood, and cognitive difficulties. Thus, sleep is essential for maintaining optimal health and well-being.&
    \phdshort&
    No&
    No&
    It seems like common knowledge that I have heard many times in the past. I have read many empirical scientific studies and listened to many professionals exclaim the many benefits of sleep hygiene over the years. This is crucial knowledge to understand in terms of vitality and longevity of our existence and mental fitness in everyday life.
    \\
    \bottomrule
    \end{tabular}
    }
    \caption{Annotations for the Explanation Informativeness User Study. The Explanation column displays the final explanation shown to participants when they viewed the explanations (see \Cref{fig:informativeness_user_evaluation} for details on the annotator’s interaction). In the last two rows, annotators entered the “need more complex explanation” path despite the explanation being at a graduate level, resulting in no available explanations and both informative and matched marked as “no.”}
    \label{tab:2.2anno_example}
\end{table*}

\subsection{Annotation Feedback Processing and Analysis}
Additionally, we processed the annotators' general feedback using \llamaseventy. For each feedback text, a function queries the LLM with a prompt that instructs it to output comma-separated labels in the format ``Aspect: direction'' (e.g., ``Vocabulary Complexity: complex'' or ``Sentence Structure: simple''). 
The query explicitly limits responses to our five predefined aspects and their valid directions (Vocabulary Complexity, Sentence Structure, Depth of Explanation, Technical Terms Usage, Overall Suitability). 
The function repeats the query if necessary (up to 10 times) until the returned labels match our predefined set. 
These labels are then used to quantify the feedback—calculating, for each aspect, the normalized difference between positive and negative mentions—which helps explain the perceived educational background in our analysis.

\begin{figure*}[ht]
    \centering
    \includegraphics[width=\textwidth]{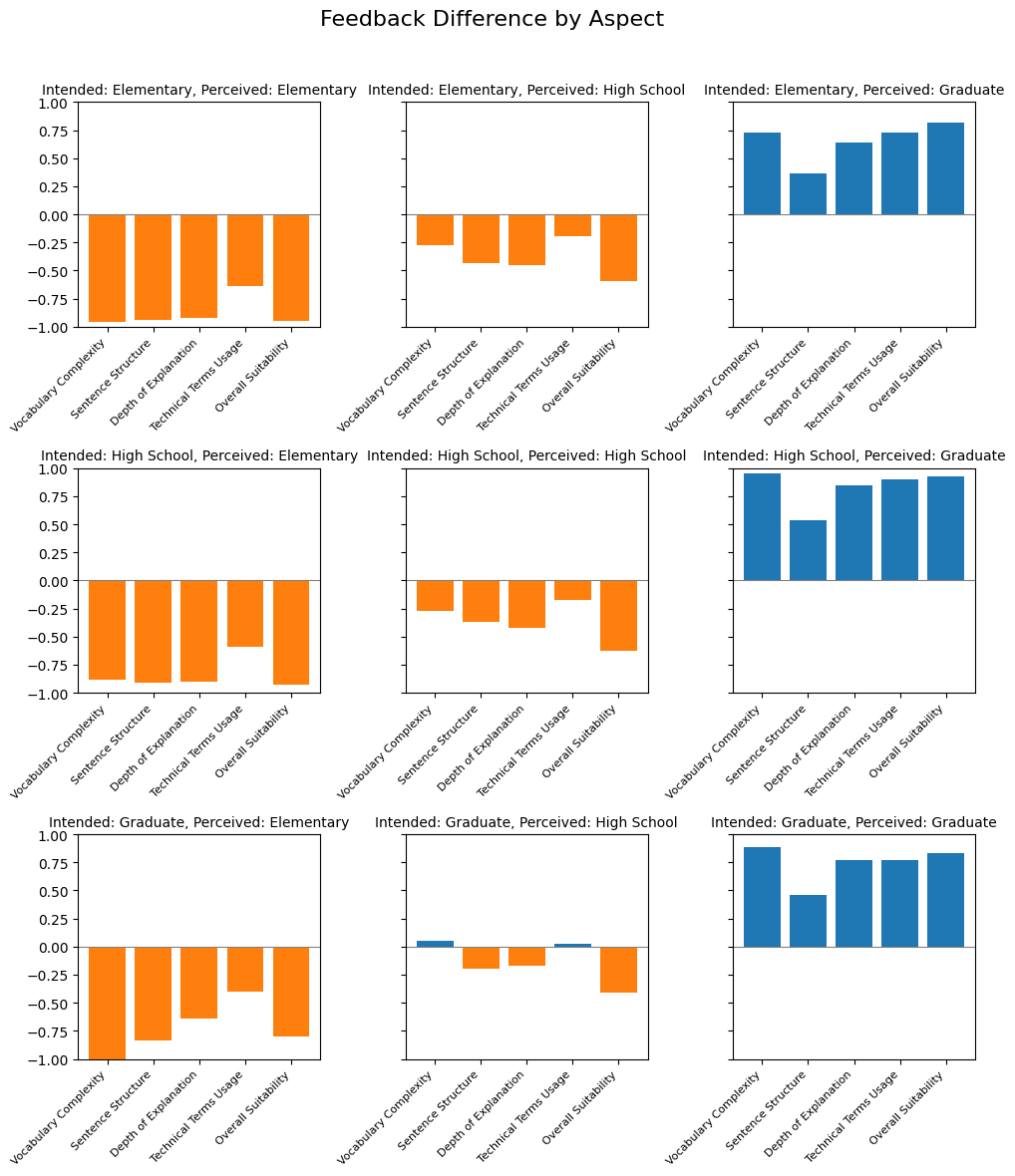}
    \caption{The normalized differences (positive minus negative counts, divided by total feedback) to our five predefined aspects (Vocabulary Complexity, Sentence Structure, Depth of Explanation, Technical Terms Usage, Overall Suitability).
    Each subplot corresponds to a specific combination of the intended (ground truth) and majority perceived educational levels (Elementary, High School, Graduate). A positive bar indicates that feedback leaned toward more complex/advanced language (e.g., ``complex,'' ``in-depth,'' ``technical''), whereas a negative bar suggests simpler characteristics.}
    \label{fig:feedback_difference}
\end{figure*}

\Cref{fig:feedback_difference} shows the normalized difference (positive minus negative counts, divided by total feedback) to our five predefined aspects.
Each subplot corresponds to a specific combination of the intended (ground truth) and majority perceived educational levels (Elementary, High School, Graduate). A positive bar indicates that feedback leaned toward more complex/advanced language (e.g., ``complex,'' ``in-depth,'' ``technical''), whereas a negative bar suggests simpler characteristics. These results align with our expectations and show that the tailored explanations exhibit distinct linguistic features corresponding to the perceived educational levels.

\section{Automated Metrics for Explanation Evaluation}
\label{sec:apx:auto_metrics}

\subsection{Metric Details}

\paragraph{Surface-form Metrics.
} Given generated explanations, we calculate the number of sentences, the average number of words per sentence, and the average reading time (which assumes 14.69 ms for each character read)
for each explanation \cite{demberg2008data} 
Additionally, we also calculate the Thing Explainer Out of Vocabulary (TE) Score proposed by \citet{august2024knowyouraudience}, which counts the ratio of words outside the top 2,000 most common words in English\footnote{This list of words was presented in the Thing Explainer book \cite{munroe2015thing} to explain scientific concepts using simple language}.

\paragraph{Readability Tests.} Given our goal of tailoring explanations to readers with varying educational backgrounds, the difficulty of the text plays a crucial role in ensuring comprehension.
Prior research in psychology has developed a variety of \textit{readability tests}, designed to quantify the difficulty of a piece of text. 
These tests typically rely on linguistic features such as sentence length, word complexity, and the presence of ``easy'' versus ``hard'' words \cite{Flesch1948-gk, o1966gobbledygook, dale1948formula}. 
Such metrics have traditionally been employed to assess the readability of textbooks, instructional materials, and technical documentation, making them highly relevant to our evaluation of explanations.

In this work, 
we leverage three widely used readability tests: Flesch-Kincaid Reading Ease, Linsear Write Formula, and Dale-Chall Readability Score. 
These were selected due to their interpretability, established validity in prior research, and diverse methodological approaches to estimating text difficulty. 
Moreover, many other readability metrics are often correlated with these three, making them representative choices for our analysis.

The \textbf{Flesch-Kincaid Reading Ease} metric assesses the readability of a text by considering the number of syllables, words, and sentences. 
Higher scores indicate text that is easier to read. 
The score is computed as:
\begin{equation} 
\small 206.835 - 1.015 \left( \frac{\text{total words}}{\text{total sentences}} \right) - 84.6 \left( \frac{\text{total syllables}}{\text{total words}} \right). 
\end{equation} 
In this formula, a higher ratio of words per sentence and syllables per word reduces the score, signaling greater difficulty. 

The \textbf{Linsear Write Formula} evaluates readability by assigning points based on word difficulty. 
``Easy words,'' with two syllables or less, earn 1 point, while ``hard words,'' with three syllables or more, earn 3 points. 
The total points are then divided by the number of sentences in the sample to produce a score \(r\). 
If \(r > 20\), the formula adjusts to \(Lw = r / 2\); otherwise, \(Lw = r / 2 - 1\). The resulting score represents the grade level. 
It emphasizes sentence structure and word complexity, providing a score that correlates with U.S. grade levels. 
Lower scores indicate text that is easier to read. 
This metric is particularly effective for identifying the complexity of short instructional or educational texts.

The \textbf{Dale-Chall Readability Score} incorporates a curated list of 3,000 common ``easy'' words. 
Words not on this list are considered ``difficult,'' and the proportion of these difficult words, combined with sentence length, determines the readability score. 
Unlike the Flesch-Kincaid metric, lower scores indicate easier text, with thresholds provided to map scores to grade levels. It is calculated as:
\begin{equation} 
\small 0.1579 \left( \frac{\text{difficult words}}{\text{words}} \times 100 \right) + 0.0496 \left( \frac{\text{words}}{\text{sentences}} \right). 
\end{equation} 
Texts with a high proportion of difficult words or longer sentences result in higher scores, reflecting increased reading difficulty.

The computed scores for all three readability metrics—Flesch-Kincaid Reading Ease, Linsear Write Formula, and Dale-Chall Readability Score—are mapped to U.S. grade levels in \Cref{tab:readability_interpretation}, which provides a detailed breakdown of score ranges and their interpreted readability levels \cite{flesch1979write, o1966gobbledygook, dale1948formula}.

\begin{table*}[h]
    \centering
    \begin{tabular}{ccl}
\toprule
\textbf{Test} & \textbf{Score Range} & \textbf{Interpretation} \\ 
\midrule
\midrule
\multirow{8}{*}{Flesch Reading Ease} 
  & 90--100  & 5th grade \\
  & 80--90   & 6th grade \\
  & 70--80   & 7th grade  \\
  & 60--70   & 8th \& 9th grade \\
  & 50--60   & 10th to 12th grade \\
  & 30--50   & College \\
  & 10--30   & College graduate \\
  & 0--10    & Professional \\
\midrule
\multirow{5}{*}{Linsear Write Formula} 
  & 0--1     & Pre-Kindergarten to 1st grade \\ 
  & 1--5     & 1st to 5th grade \\ 
  & 5--8     & 5th to 8th grade \\ 
  & 8--11    & 8th to 11th grade \\ 
  & 11+      & 11th grade to college \\
\midrule
\multirow{6}{*}{Dale-Chall Readability} 
  & 0--5     & 4th grade or lower \\ 
  & 5--6     & 5th--6th grade level \\ 
  & 6--7     & 7th--8th grade level \\ 
  & 7--8     & 9th--10th grade level \\ 
  & 8--9     & 11th--12th grade level \\ 
  & 9--10    & College level \\ 
\bottomrule
\end{tabular}
    \caption{Readability Score Interpretations}
    \label{tab:readability_interpretation}
\end{table*}

\paragraph{Reasoning Type. } 
In addition to surface form metrics and readability tests, we also characterize the kind of reasoning used to answer these ``why'' questions.
``Why'' questions can be answered in multiple ways, but the two predominant manners are mechanistic (explaining the \textit{process} or \textit{mechanism} of how a phenomenon came to be) or teleological (explaining the \textit{purpose} or \textit{function} of the phenomenon) \cite{lombrozo2014explanation}, and this reasoning style has often been shown to distinguish explanations preferred by people from various educational backgrounds like children or students \cite{mccarthy2023right, kelemen1999rocks}.
\Eg, ``Why is the sky blue?'' can be answered either by explaining rayleigh scattering (mechanistic) or that the blue sky has a prettier contrast (teleological).

\begin{table}[h]
    \centering
    \begin{tabular}{cl}
        \toprule
        \textbf{Config} & \textbf{Assignment} \\
        \midrule
        \multirow{2}{*}{model} & \textbf{\llama-4bit} \\
        & Number of parameters: 3 billion \\
        \midrule
        \midrule
        max\_tokens & 10 \\
        temperature & 0.1 \\
        GPU & Apple M4 Max \\
        Inferring time & 1 hour per 13k explanations\\
        \midrule
        \midrule        
        \multicolumn{2}{p{0.48\textwidth}}{\textbf{Prompt:} You are a helpful assistant. Below is a question followed by an explanation.
        For the explanation, classify whether it is mechanistic or teleological.
        A mechanistic explanation describes how something happens, focusing on the processes, systems, or mechanisms involved.
        A teleological explanation describes something in terms of its goal, purpose, or intended outcome.
        
        Question: [Question]
        
        Explanation: [Explanation]
        
        Please provide your classification in one word: mechanistic or teleological.
        }\\
        \bottomrule
    \end{tabular}
    \caption{Model Configuration and Prompting Details for Reasoning Type Classification.}
    \label{tab:mech_teleo_classification_model_config}
\end{table}

For this classification task, we employ \llama using the prompt and configurations in \Cref{tab:mech_teleo_classification_model_config}.

\subsection{Significance tests in automated metric evaluation}
\label{sec:apx:metric_significance_tests}

Beyond average values, we also examine the distribution of scores for each metric (see \Cref{sec:apx:auto_score_dist}), and observe significant differences in the scores.
We evaluate two aspects: (1) a two-tailed test assessing whether explanation distributions are significantly distinct across these groups (\sigoverall) and (2) a one-sided test verifying whether metric values follow the expected complexity ordering (\sigorder), i.e., \es < \hs < \phd. 
In both tests, across all explanation pairs, we observe statistically significant differences ($p < 0.001$), confirming that \gptfour's explanations not only vary in complexity but also follow a systematic progression in difficulty.
We conduct significance tests for all three pairs of explanation types (\es vs. \hs, \hs vs. \phd, and \es vs. \phd) using the Kolmogorov-Smirnov (KS) test. 
We opted for the KS test because it is a nonparametric method that compares entire distributions without assuming normality. 
Given that many of our distributions are not bell-shaped, a t-test which assumes normality would not be suitable here.

\subsection{Correlation between automated metrics and user perceived educational backgrounds}
\label{sec:apx:corr_ratio}

We computed the \emph{correlation ratio} across different automatic readability metrics to assess the correlation of annotator judgments with automated metrics. The correlation ratio measures the categorial-continuous association \cite{fisher1970statistical}. Higher values indicate stronger association between the automated readability scores and the perceived educational background.

\Cref{tab:correlation_ratio} summarize the correlation ratios for various readability metrics. Notably, the Flesch Reading Ease score shows a medium correlation (0.425) with the annotator-perceived educational background, indicating that automated assessments align well with human judgments.

\begin{table}[H] 
    \small
    \centering 
    \begin{tabular}{lc} 
    \toprule 
    \textbf{Metric} & \textbf{Correlation Ratio $\eta$} \\ 
    \midrule Num Sentences & 0.360 \\ 
    Avg Words Per Sentence & 0.177 \\
    Reading Time & 0.511 \\
    TE Score & 0.441 \\
    Flesch Reading Ease & 0.425 \\ 
    Linsear Write Formula & 0.291 \\
    Dale Chall Readability Score& 0.324 \\
    \bottomrule 
    \end{tabular} 
    \caption{Correlation ratios between automated readability metrics and annotator-perceived educational background.} \label{tab:correlation_ratio} 
\end{table}

\subsection{Readability and Explanation Type Distributions}
\label{sec:apx:auto_score_dist}

We report the distributions of several automated metrics computed over the generated explanations.
These metrics include surface-form attributes such as sentence count, words per sentence, and estimated reading time, as well as traditional readability tests including Flesch-Kincaid Reading Ease, Linsear Write Formula, and Dale-Chall Readability Score.

\Cref{tab:automated metrics_apx} presents the aggregated scores across all explanations, while \Cref{tab:automated metrics stem} and \Cref{tab:automated metrics nonstem} provide a breakdown for STEM and non-STEM questions, respectively.

\begin{figure*}[t]
    \centering
    \includegraphics[width=\textwidth]{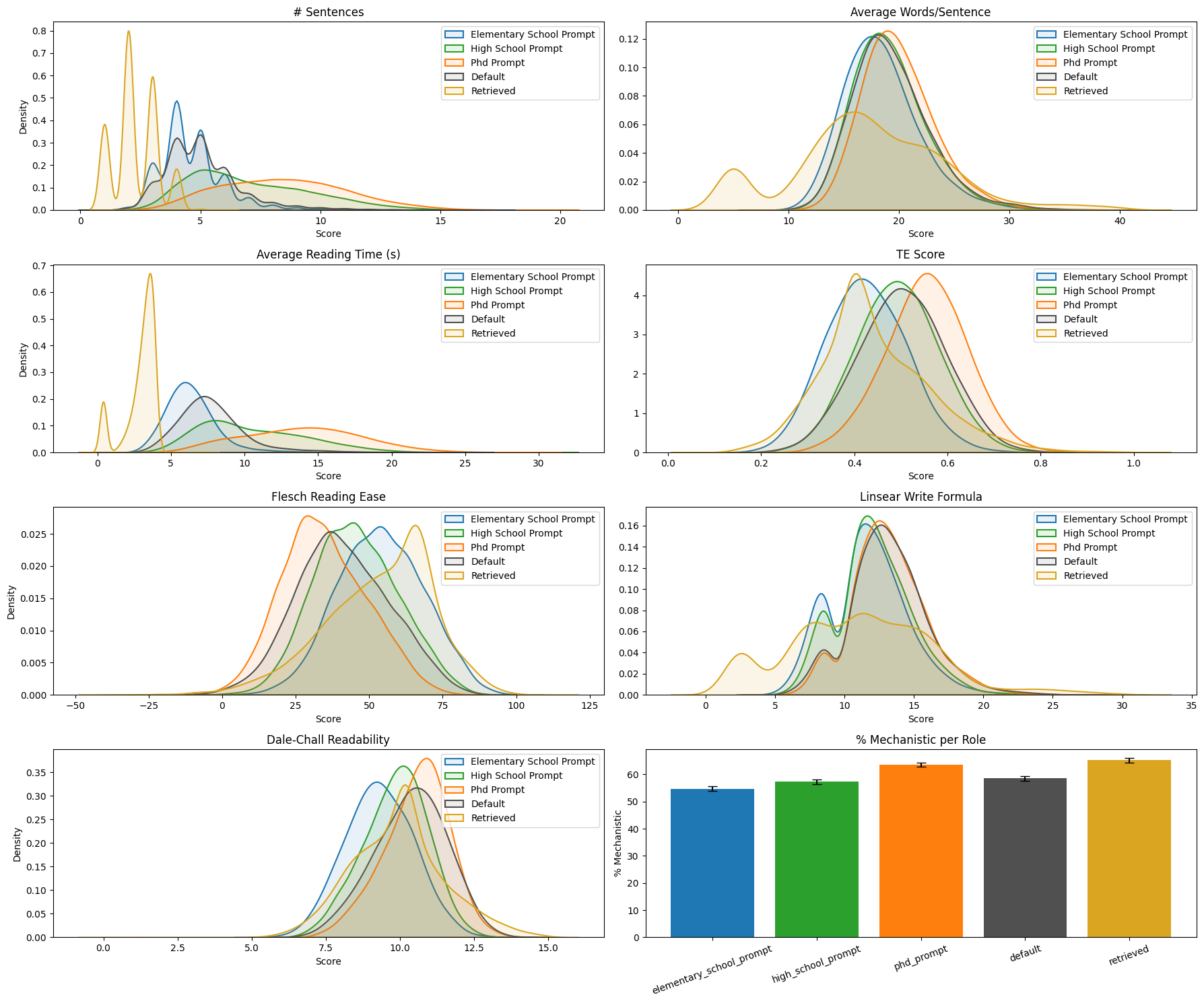}
    \caption{Overall distribution of readability scores with explanations generated by \gptfour}
    \label{fig:auto_metrics_distributions}
\end{figure*}

\Cref{fig:auto_metrics_distributions} shows the overall distribution of readability scores with explanations generated by \gptfour, with trends consistent with our expectations.
We observe that explanations tailored for audiences with higher educational backgrounds exhibit longer sentences and increased reading times, leading to lower Flesch-Kincaid scores and higher Linsear Write and Dale-Chall scores.

\begin{figure*}[t]
    \centering
    \includegraphics[width=\textwidth]{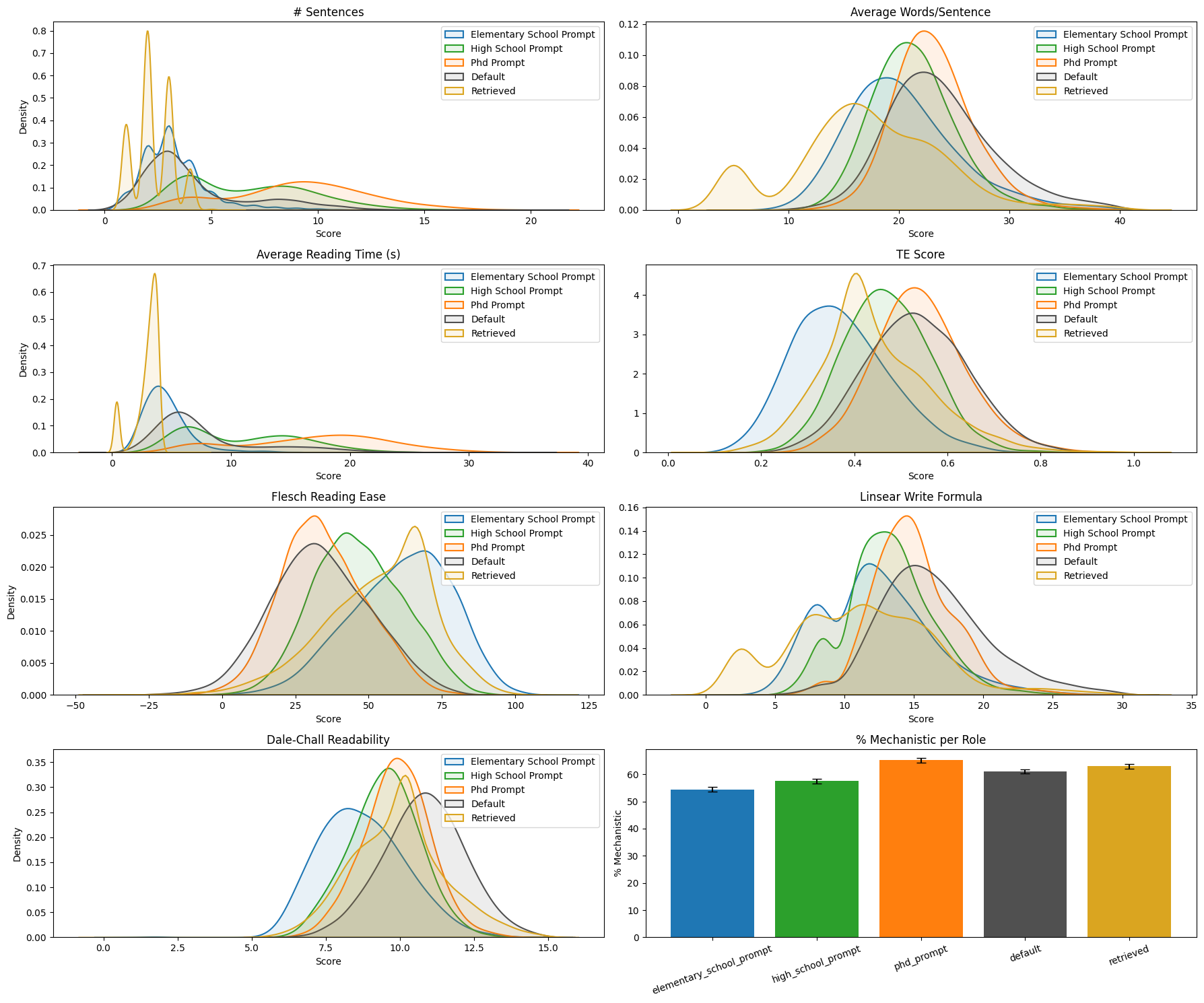}
    \caption{Overall distribution of readability scores with explanations generated by \llama.}
    \label{fig:auto_metrics_distributions_llama}
\end{figure*}

\Cref{fig:auto_metrics_distributions_llama} displays the readability distribution evaluated using \llama. The overall trend is the same.
In rare cases (<1\%) the model generated repetitive phrases (e.g., ``gentrification-induced gentrification-induced gentrification-induced...''), which significantly impacted the variance of certain readability metrics (up to 50\%).
To mitigate this effect, we applied a rough filtering step to remove these anomalous outputs prior to analysis.

\begin{figure*}[t]
    \centering
    \includegraphics[width=\textwidth]{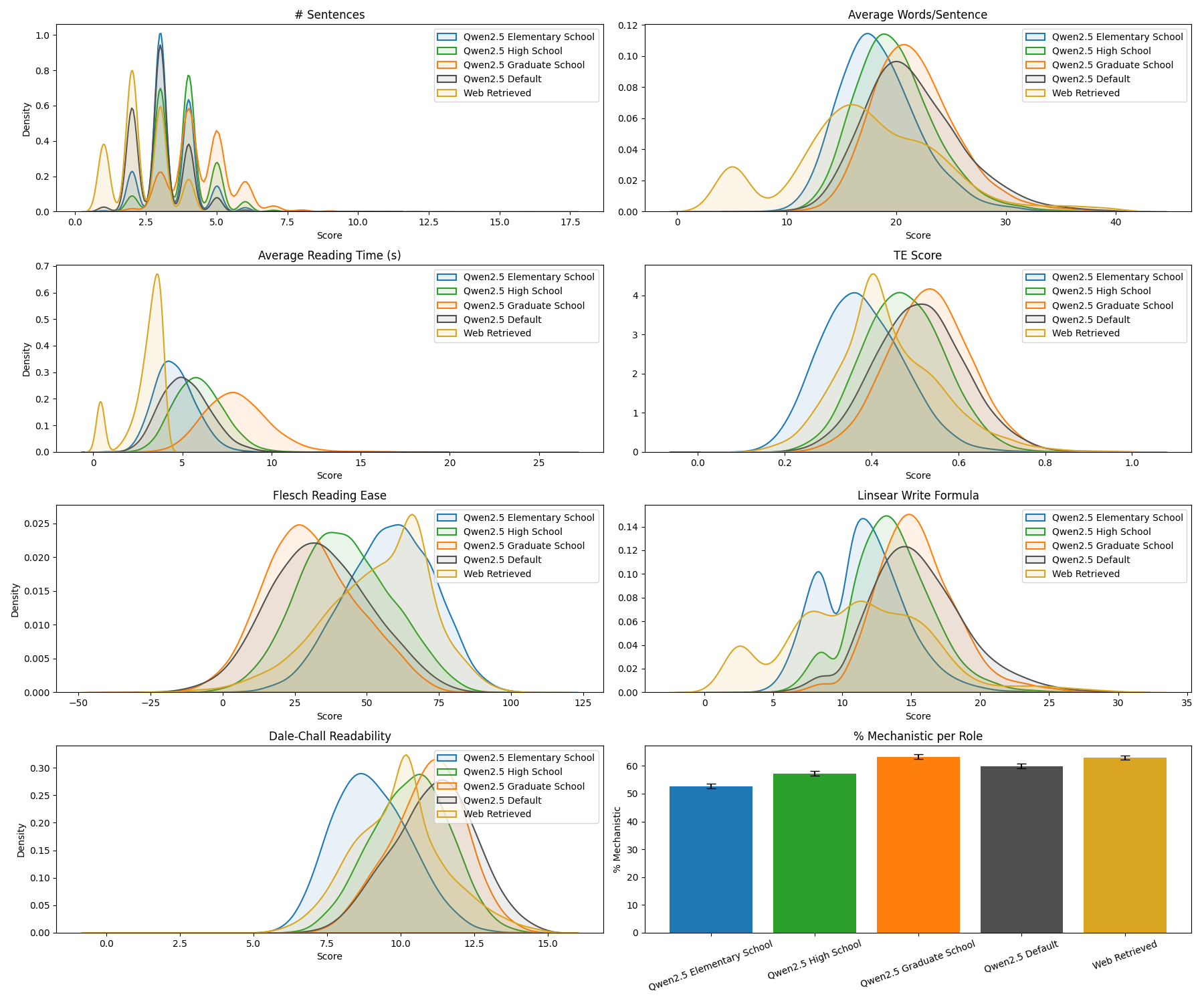}
    \caption{Overall distribution of readability scores with explanations generated by \qwen.}
    \label{fig:auto_metrics_distributions_qwen}
\end{figure*}

\begin{figure*}[t]
    \centering
    \includegraphics[width=\textwidth]{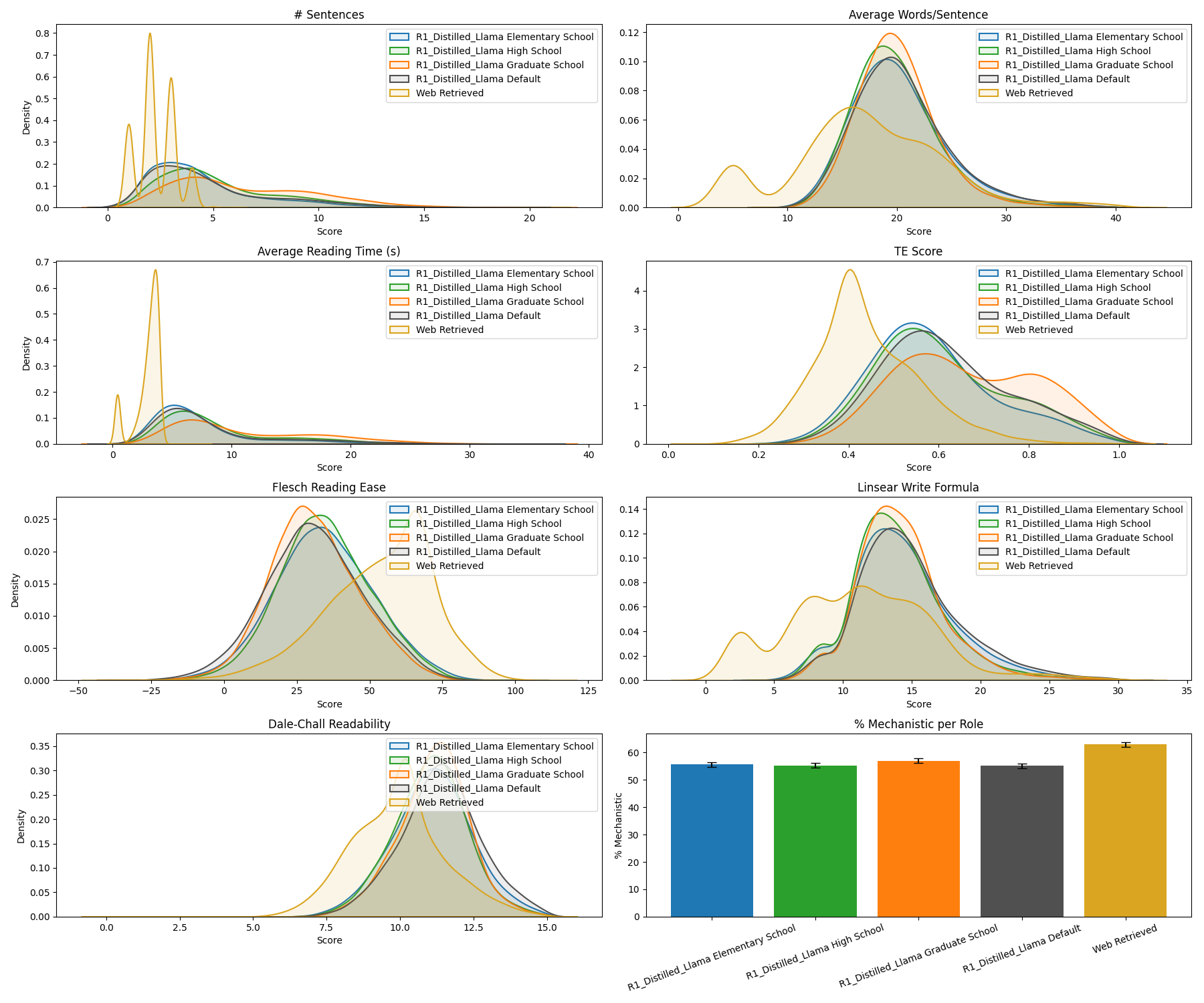}
    \caption{Overall distribution of readability scores with explanations generated by \deepseek.}
    \label{fig:auto_metrics_distributions_deepseek}
\end{figure*}

\Cref{fig:auto_metrics_distributions_qwen} and \Cref{fig:auto_metrics_distributions_deepseek} displaus readability distributions for \qwen and \deepseek respectively.

When splitting the data into STEM (\Cref{tab:automated metrics stem}) and non-STEM (\Cref{tab:automated metrics nonstem}) domains, we observe the same overall trends in readability across educational levels.
However, the reasoning type distribution shows a divergence between domains.
For STEM questions, the mechanistic reasoning proportion is significantly higher, approaching 90\% for both \gptfour and \llama models' explanations, whereas for non-STEM questions, the proportion for mechanistic reasoning is generally around 30\%.

\begin{table*}[!t]
\centering
\resizebox{\linewidth}{!}{
\begin{tabular}{llccccccccc}
\toprule
\textbf{Type of Metric} & \textbf{Metric} & \multicolumn{4}{c}{\gptfour} & \multicolumn{4}{c}{\llama}  \\
\cmidrule(lr){3-6} \cmidrule(lr){7-10}
 & & \es & \hs & \phd & \default &\es & \hs & \phd & \default & \google \\ 
\midrule
\midrule 
\multirow{4}{*}{\parbox{0.8cm}{Surface-form}} 
& \# Sentences             
  & 04.63 \(\pm\) \small{01.34}
  & 07.08 \(\pm\) \small{02.53}
  & 08.46 \(\pm\) \small{02.62}
  & 05.07 \(\pm\) \small{01.63} 
  
  & 03.29 \(\pm\) \small{01.63}
  & 06.70 \(\pm\) \small{02.97}
  & 09.10 \(\pm\) \small{03.33}
  & 04.24 \(\pm\) \small{02.63}
  & 02.30 \(\pm\) \small{00.90} \\
& Avg. \# Words / Sentence 
  & 18.43 \(\pm\) \small{03.47}
  & 19.17 \(\pm\) \small{03.36}
  & 20.00 \(\pm\) \small{03.38}
  & 19.35 \(\pm\) \small{03.57}
  
  & 20.39 \(\pm\) \small{05.10}
  & 21.30 \(\pm\) \small{03.81}
  & 23.12 \(\pm\) \small{03.69}
  & 23.74 \(\pm\) \small{04.88}
  & 17.26 \(\pm\) \small{06.90} \\
& Avg. Reading Time (s)     
  & 06.36 \(\pm\) \small{01.75}
  & 10.57 \(\pm\) \small{03.65}
  & 13.93 \(\pm\) \small{04.05}
  & 07.81 \(\pm\) \small{02.41}
  
  & 04.61 \(\pm\) \small{02.14}
  & 10.97 \(\pm\) \small{05.00}
  & 17.05 \(\pm\) \small{06.30}
  & 07.93 \(\pm\) \small{04.71}
  & 02.93 \(\pm\) \small{01.04} \\
& TE Score                  
  & 00.43 \(\pm\) \small{00.09}
  & 00.49 \(\pm\) \small{00.09}
  & 00.55 \(\pm\) \small{00.09}
  & 00.50 \(\pm\) \small{00.09}
  
  & 00.37 \(\pm\) \small{00.11}
  & 00.47 \(\pm\) \small{00.09}
  & 00.54 \(\pm\) \small{00.10}
  & 00.53 \(\pm\) \small{00.11}
  & 00.44 \(\pm\) \small{00.12} \\
\midrule
\multirow{3}{*}{\parbox{0.8cm}{Readability Tests}} 
& Flesch-Kincaid Reading Ease ($\downarrow$)
  & 53.82 \(\pm\) \small{14.52}
  & 45.51 \(\pm\) \small{14.26}
  & 34.70 \(\pm\) \small{14.43}
  & 41.00 \(\pm\) \small{15.73}
  
  & 60.91 \(\pm\) \small{17.14}
  & 46.39 \(\pm\) \small{15.11}
  & 34.56 \(\pm\) \small{14.51}
  & 33.68 \(\pm\) \small{16.87}
  & 53.35 \(\pm\) \small{18.52} \\
& Linsear Write Formula ($\uparrow$)     
  & 11.67 \(\pm\) \small{02.77}
  & 12.15 \(\pm\) \small{02.70}
  & 13.21 \(\pm\) \small{02.70}
  & 13.16 \(\pm\) \small{02.82}
  
  & 12.29 \(\pm\) \small{03.94}
  & 13.23 \(\pm\) \small{03.09}
  & 14.88 \(\pm\) \small{02.93}
  & 16.43 \(\pm\) \small{03.93}
  & 11.10 \(\pm\) \small{05.21} \\
& Dale-Chall Readability Score ($\uparrow$)
  & 09.31 \(\pm\) \small{01.16}
  & 09.84 \(\pm\) \small{01.10}
  & 10.55 \(\pm\) \small{01.10}
  & 10.35 \(\pm\) \small{01.23} 
  
  & 08.73 \(\pm\) \small{01.47}
  & 09.49 \(\pm\) \small{01.19}
  & 09.88 \(\pm\) \small{01.13}
  & 10.80 \(\pm\) \small{01.38}
  & 09.94 \(\pm\) \small{01.56} \\
\hline
Reasoning Type 
& \% Mechanistic Reasoning               
  & 54.76\% & 57.28\% & 63.54\% & 58.51\% & 55.04\% & 57.81\% & 65.65\% & 63.03\% & 65.17\%\\
\bottomrule
\end{tabular}}
\caption{Comparison of surface-form, readability, and reasoning-type metrics across different education levels for \gptfour and \llama, along with retrieved explanations.}
\label{tab:automated metrics_apx}
\end{table*}

\begin{table*}[!t]
\centering
\resizebox{\linewidth}{!}{
\begin{tabular}{llccccccccc}
\toprule
\textbf{Type of Metric} & \textbf{Metric} & \multicolumn{4}{c}{\gptfour} & \multicolumn{4}{c}{\llama}  \\
\cmidrule(lr){3-6} \cmidrule(lr){7-10}
 & & \es & \hs & \phd & \default &\es & \hs & \phd & \default & \google \\ 
\midrule
\midrule 
\multirow{4}{*}{\parbox{0.8cm}{Surface-form}} 
& \# Sentences             
  & 04.39 \(\pm\) \small{01.28}
  & 06.19 \(\pm\) \small{02.12}
  & 07.44 \(\pm\) \small{02.37}
  & 04.65 \(\pm\) \small{01.43}
  
  & 03.24 \(\pm\) \small{01.54}
  & 05.68 \(\pm\) \small{02.55}
  & 07.64 \(\pm\) \small{03.14}
  & 03.39 \(\pm\) \small{01.81}
  & 02.52 \(\pm\) \small{00.86} \\
& Avg. \# Words / Sentence 
  & 18.82 \(\pm\) \small{03.78}
  & 19.83 \(\pm\) \small{03.62}
  & 20.87 \(\pm\) \small{03.58}
  & 19.88 \(\pm\) \small{03.90}
  
  & 19.56 \(\pm\) \small{05.24}
  & 20.80 \(\pm\) \small{04.11}
  & 23.09 \(\pm\) \small{04.11}
  & 23.97 \(\pm\) \small{05.22}
  & 17.86 \(\pm\) \small{05.96} \\
& Avg. Reading Time (s)     
  & 05.88 \(\pm\) \small{01.70}
  & 09.10 \(\pm\) \small{03.10}
  & 12.14 \(\pm\) \small{03.75}
  & 06.97 \(\pm\) \small{02.11}
  
  & 04.14 \(\pm\) \small{01.68}
  & 08.57 \(\pm\) \small{03.96}
  & 13.51 \(\pm\) \small{05.63}
  & 06.03 \(\pm\) \small{02.94}
  & 03.12 \(\pm\) \small{00.79} \\
& TE Score                  
  & 00.42 \(\pm\) \small{00.09}
  & 00.46 \(\pm\) \small{00.09}
  & 00.52 \(\pm\) \small{00.09}
  & 00.47 \(\pm\) \small{00.10}
  
  & 00.37 \(\pm\) \small{00.10}
  & 00.45 \(\pm\) \small{00.10}
  & 00.51 \(\pm\) \small{00.11}
  & 00.49 \(\pm\) \small{00.11}
  & 00.44 \(\pm\) \small{00.12} \\
\midrule
\multirow{3}{*}{\parbox{0.8cm}{Readability Tests}} 
& Flesch-Kincaid Reading Ease ($\downarrow$)
  & 60.61 \(\pm\) \small{13.82}
  & 53.30 \(\pm\) \small{13.47}
  & 42.89 \(\pm\) \small{13.76}
  & 49.41 \(\pm\) \small{14.99}
  
  & 66.97 \(\pm\) \small{15.60}
  & 55.03 \(\pm\) \small{14.10}
  & 42.81 \(\pm\) \small{14.10}
  & 42.48 \(\pm\) \small{16.57}
  & 57.61 \(\pm\) \small{17.38} \\
& Linsear Write Formula ($\uparrow$)     
  & 11.40 \(\pm\) \small{03.00}
  & 12.10 \(\pm\) \small{02.84}
  & 13.16 \(\pm\) \small{02.87}
  & 12.91 \(\pm\) \small{03.09}
  
  & 11.39 \(\pm\) \small{04.02}
  & 12.57 \(\pm\) \small{03.25}
  & 14.43 \(\pm\) \small{03.23}
  & 16.09 \(\pm\) \small{04.18}
  & 11.13 \(\pm\) \small{04.50} \\
& Dale-Chall Readability Score ($\uparrow$)
  & 08.81 \(\pm\) \small{01.10}
  & 09.23 \(\pm\) \small{01.04}
  & 09.92 \(\pm\) \small{01.08}
  & 09.69 \(\pm\) \small{01.17}
  
  & 08.39 \(\pm\) \small{01.36}
  & 08.93 \(\pm\) \small{01.17}
  & 09.39 \(\pm\) \small{01.17}
  & 10.22 \(\pm\) \small{01.41}
  & 09.54 \(\pm\) \small{01.52} \\
\hline
Reasoning Type 
& \% Mechanistic Reasoning               
  & 88.49\% & 89.91\% & 93.09\% & 91.58\% & 89.26\% & 92.59\% & 95.26\% & 94.40\% & 93.92\%\\
\bottomrule
\end{tabular}}
\caption{Comparison of surface-form, readability, and reasoning-type metrics across different education levels for \gptfour and \llama, on the STEM questions split.}
\label{tab:automated metrics stem}
\end{table*}

\begin{table*}[!t]
\centering
\resizebox{\linewidth}{!}{
\begin{tabular}{llccccccccc}
\toprule
\textbf{Type of Metric} & \textbf{Metric} & \multicolumn{4}{c}{\gptfour} & \multicolumn{4}{c}{\llama}  \\
\cmidrule(lr){3-6} \cmidrule(lr){7-10}
 & & \es & \hs & \phd & \default &\es & \hs & \phd & \default & \google \\ 
\midrule
\midrule 
\multirow{4}{*}{\parbox{0.8cm}{Surface-form}} 
& \# Sentences             
  & 04.83 \(\pm\) \small{01.36}
  & 07.85 \(\pm\) \small{02.61}
  & 09.35 \(\pm\) \small{02.51}
  & 05.44 \(\pm\) \small{01.70}
  
  & 03.34 \(\pm\) \small{01.70}
  & 07.59 \(\pm\) \small{03.03}
  & 10.36 \(\pm\) \small{02.96}
  & 04.97 \(\pm\) \small{02.98}
  & 02.12 \(\pm\) \small{00.88} \\
& Avg. \# Words / Sentence 
  & 18.09 \(\pm\) \small{03.13}
  & 18.60 \(\pm\) \small{03.01}
  & 19.25 \(\pm\) \small{02.99}
  & 18.88 \(\pm\) \small{03.18}
  
  & 21.10 \(\pm\) \small{04.86}
  & 21.72 \(\pm\) \small{03.47}
  & 23.15 \(\pm\) \small{03.30}
  & 23.55 \(\pm\) \small{04.55}
  & 16.72 \(\pm\) \small{07.60} \\
& Avg. Reading Time (s)     
  & 06.77 \(\pm\) \small{01.69}
  & 11.84 \(\pm\) \small{03.62}
  & 15.49 \(\pm\) \small{03.64}
  & 08.53 \(\pm\) \small{02.42}
  
  & 05.02 \(\pm\) \small{02.39}
  & 13.05 \(\pm\) \small{04.88}
  & 20.10 \(\pm\) \small{05.14}
  & 09.57 \(\pm\) \small{05.29}
  & 02.77 \(\pm\) \small{01.20} \\
& TE Score                  
  & 00.44 \(\pm\) \small{00.09}
  & 00.52 \(\pm\) \small{00.08}
  & 00.58 \(\pm\) \small{00.07}
  & 00.53 \(\pm\) \small{00.08}
  
  & 00.37 \(\pm\) \small{00.11}
  & 00.49 \(\pm\) \small{00.09}
  & 00.56 \(\pm\) \small{00.09}
  & 00.57 \(\pm\) \small{00.10}
  & 00.45 \(\pm\) \small{00.12} \\
\midrule
\multirow{3}{*}{\parbox{0.8cm}{Readability Tests}} 
& Flesch-Kincaid Reading Ease ($\downarrow$)
  & 47.93 \(\pm\) \small{12.38}
  & 38.75 \(\pm\) \small{11.15}
  & 27.62 \(\pm\) \small{10.80}
  & 33.73 \(\pm\) \small{12.38}
  
  & 55.63 \(\pm\) \small{16.67}
  & 38.89 \(\pm\) \small{11.53}
  & 27.44 \(\pm\) \small{10.55}
  & 26.12 \(\pm\) \small{13.01}
  & 49.60 \(\pm\) \small{18.69} \\
& Linsear Write Formula ($\uparrow$)     
  & 11.91 \(\pm\) \small{02.54}
  & 12.19 \(\pm\) \small{02.57}
  & 13.26 \(\pm\) \small{02.55}
  & 13.37 \(\pm\) \small{02.53}
  
  & 13.07 \(\pm\) \small{03.69}
  & 13.81 \(\pm\) \small{02.83}
  & 15.27 \(\pm\) \small{02.58}
  & 16.72 \(\pm\) \small{03.67}
  & 11.07 \(\pm\) \small{05.76} \\
& Dale-Chall Readability Score ($\uparrow$)
  & 09.74 \(\pm\) \small{01.03}
  & 10.38 \(\pm\) \small{00.83}
  & 11.09 \(\pm\) \small{00.79}
  & 10.93 \(\pm\) \small{00.96} 
  
  & 09.03 \(\pm\) \small{01.50}
  & 09.97 \(\pm\) \small{00.97}
  & 10.31 \(\pm\) \small{00.90}
  & 11.31 \(\pm\) \small{01.12}
  & 10.29 \(\pm\) \small{01.50} \\
\hline
Reasoning Type 
& \% Mechanistic Reasoning               
  & 25.55\% & 29.02\% & 37.97\% & 29.88\% & 25.26\% & 27.63\% & 40.05\% & 36.05\% & 39.92\%\\
\bottomrule
\end{tabular}}
\caption{Comparison of surface-form, readability, and reasoning-type metrics across different education levels for \gptfour and \llama, on the Non-STEM questions split.}
\label{tab:automated metrics nonstem}
\end{table*}

\subsection{Informational overlap between tailored and baseline explanations}
\label{sec:apx:similarity}

To analyze the informational overlap between explanations, we introduce the TE Set Difference (\tesdiff).
While the TE Score captures the proportion of simple, commonly used words in an explanation, \tesdiff measures \textit{how much} of the TE word set from one explanation \textit{is missing in another}. 
Specifically, it quantifies the proportion of TE words in one explanation that do not appear in another, normalized by the total number of unique TE words in the first explanation:
\begin{equation}
    \text{\tesdiff} (x,y) = \frac{|(x\backslash y)\cap TE|}{| x \cap TE|}
\end{equation}
A higher \tesdiff score indicates that the explanation $x$ contains significantly
more unique information compared to $y$, whereas a lower \tesdiff score suggests greater overlap between the two explanations, meaning they are more similar in terms of TE word usage.

\begin{figure}[H]
    \centering
    \includegraphics[width=\linewidth]{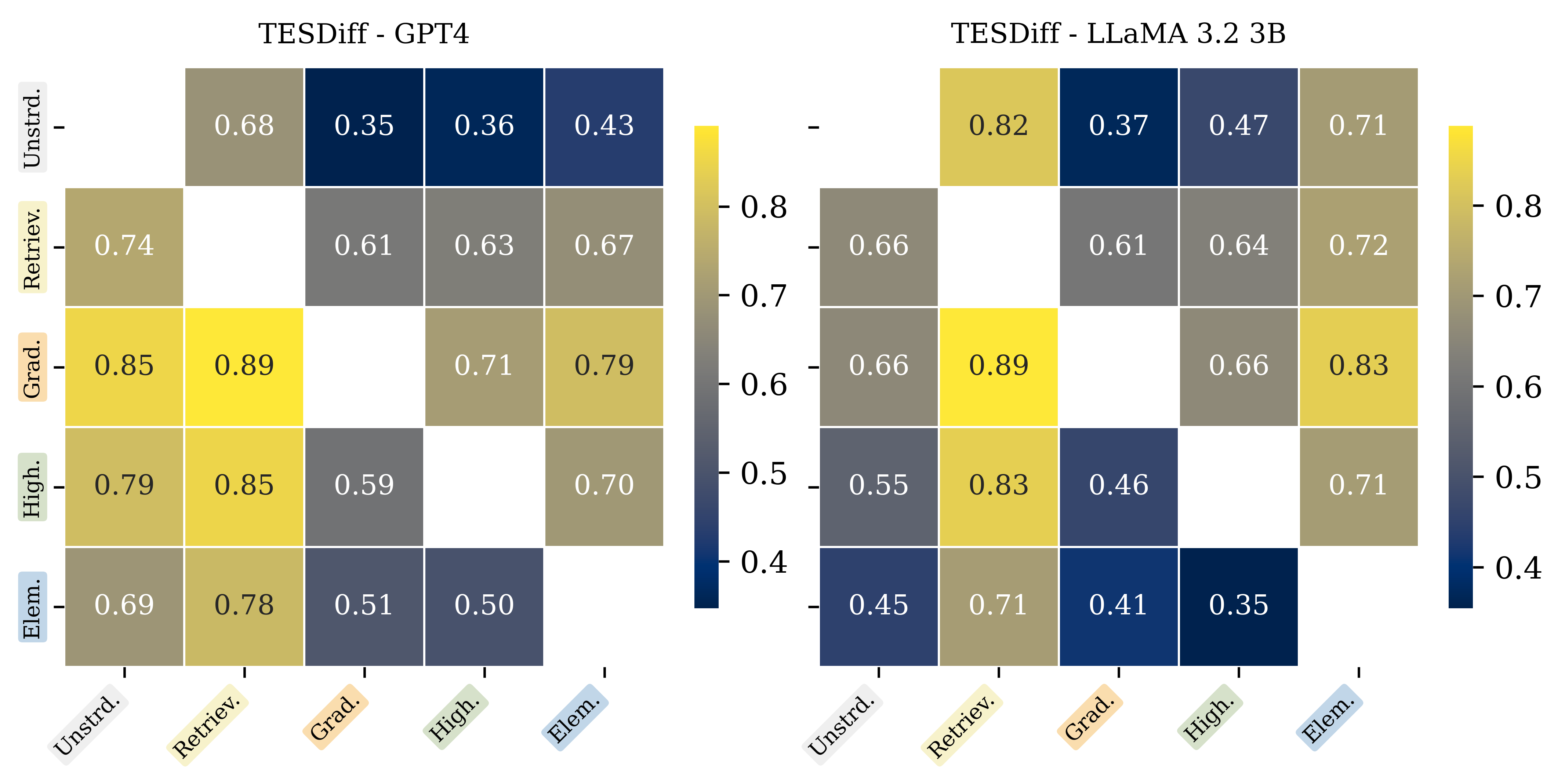}
    \caption{Heatmap of TESDiff scores across different explanation types for \gptfour (left) and \llama (right). Higher TESDiff scores indicate greater divergence from general-purpose responses, with ``Graduate School'' explanations consistently exhibiting the highest TESDiff values across models.}
    \label{fig:tesdiff_heatmap}
\end{figure}

\Cref{fig:tesdiff_heatmap} presents a heatmap of \tesdiff scores across different explanation types
We observe that explanations generated for \phd have consistently \textbf{high} \tesdiff scores when compared to other explanations, suggesting that they introduce the most new information and diverge the most from general-purpose responses. 
In contrast, explanations generated for \hs and \es audiences exhibit lower \tesdiff scores between them, indicating that they are more similar to each other and that \gptfour does not make as strong a distinction between these two levels.

While looking at \default and other explanations, we find that \es and \hs prompts produce the most similar explanations to the \default setting, as indicated by their lower \tesdiff scores when compared to default explanations. 
This suggests that when \gptfour generates explanations \textit{without} an explicitly assigned educational background,
the resulting explanations tend to align more with these middle-ground audiences. 
On the other hand, \google explanations show greater similarity to \hs and \phd explanations, likely because web sources often contain more technical detail than \default responses.

Automated metrics for evaluating explanation complexity and similarity indicate that \gptfour generates explanations with distinct scores for different audience roles.
Surface-form attributes, readability tests, and the \tesdiff score all show systematic differences across educational levels, suggesting that explanations vary in length, complexity, and word choice. 
However, these numerical differences do not necessarily imply that the explanations are \textit{useful or well-suited for their intended audiences}. 
While \gptfour can produce explanations that differ in measurable ways, whether these differences actually improve comprehension or align with user needs requires further human evaluation, which we explore in the Human Experiment section.

\section{Case study with ELI5 subreddit questions}
\label{sec:apx:case_study_eli5}
To examine why explanation complexity varies, we conducted a case study using the \textit{Explain Like I'm Five} (ELI5) subreddit. 
For each question where we performed human annotations on explanation readability, we retrieved the most similar question from ELI5 using the Reddit API. 
We queried up to 25 relevant posts and identified the closest match using the SentenceTransformer \texttt{all-MiniLM-L6-v2}.

\Cref{tab:eli5_match} presents a selection of our findings, including the original question, its most similar counterpart in ELI5, the similarity score, and the intended-to-perceived educational mapping. 
Our analysis reveals a consistent trend: when a question has a highly similar counterpart in ELI5, explanations tend to be oversimplified, particularly for prompts originally intended for higher educational levels. 
This suggests that when tasked with explaining such questions, the model is more likely to produce simpler explanations, potentially due to their prevalence in training data.

\begin{table*}[t]
    \centering
    \resizebox{\linewidth}{!}{
    \begin{tabular}{p{0.16\textwidth} p{0.16\textwidth} p{0.2\textwidth} p{0.1\textwidth} p{0.32\textwidth}}
        \toprule
        \textbf{Question} & \textbf{Best Match Reddit ELI5 Question} & \textbf{Reddit ELI5 link} & \textbf{Similarity Score} & \textbf{Intended -> Perceived Mapping}\\
        \midrule
        Why do carnivorous plants eat insects?	 & Why do carnivorous plants eat meat? 
        & \url{https://www.reddit.com/r/explainlikeimfive/comments/trx3rw/eli5_why_do_carnivorous_plants_eat_meat/} 
        & 0.83 
        & \parbox[t]{0.32\textwidth}{\raggedright \esshort -> \esshort,\\ \hsshort -> \hsshort,\\ \phdshort -> \esshort}\\

        Why do we perceive different colors?
        & Why do we see colours other than red, green and blue?
        & \url{https://www.reddit.com/r/explainlikeimfive/comments/rihdyj/eli5_why_do_we_see_colours_other_than_red_green/}
        & 0.85
        & \parbox[t]{0.32\textwidth}{\raggedright \esshort -> \esshort,\\ \hsshort -> \esshort,\\ \phdshort -> \esshort}\\

        Why do balloons pop when exposed to sharp objects?
        & Why do bubbles pop when otherwise undisturbed?
        & \url{https://www.reddit.com/r/explainlikeimfive/comments/p7kowq/eli5_why_do_bubbles_pop_when_otherwise_undisturbed/}
        & 0.55
        & \parbox[t]{0.32\textwidth}{\raggedright \esshort -> \esshort,\\ \hsshort -> \hsshort,\\ \phdshort -> \hsshort}\\

        Why is it beneficial for an organism to adapt?
        & Why would viruses and bacteria ever try to harm their host?
        & \url{https://www.reddit.com/r/explainlikeimfive/comments/1optwy/eli5_why_would_viruses_and_bacteria_ever_try_to/}
        & 0.43
        & \parbox[t]{0.32\textwidth}{\raggedright \esshort -> \hsshort,\\ \hsshort -> \hsshort,\\ \phdshort -> \hsshort}\\

        Why do we mourn the loss of celebrity we never met?
        & What does this poem mean?
        & \url{https://www.reddit.com/r/explainlikeimfive/comments/1902oc/what_does_this_poem_mean/}
        & 0.16
        & \parbox[t]{0.32\textwidth}{\raggedright \esshort -> \esshort,\\ \hsshort -> \hsshort,\\ \phdshort -> \phdshort}\\

        Why have ``perennial topics" like love, death, and identity remained key themes in literature?
        & Why is Death Valley one of the hottest places on earth despite being far from the equator?
        & \url{https://www.reddit.com/r/explainlikeimfive/comments/1hfcnvl/eli5_why_is_death_valley_one_of_the_hottest/}
        &
        0.12
        & \parbox[t]{0.32\textwidth}{\raggedright \esshort -> \hsshort,\\ \hsshort -> \hsshort,\\ \phdshort -> \phdshort}\\
        \bottomrule
    \end{tabular}
    }
    \caption{Examples of ELI5 question matching with similarity scores. When a question has a highly similar variant in the ELI5 subreddit, there is a tendency for oversimplification for prompts originally intended for higher educational levels.}

    \label{tab:eli5_match}
\end{table*}

\end{document}